\documentclass{article}

\usepackage{arxiv}

\usepackage[utf8]{inputenc} 
\usepackage[T1]{fontenc}    
\usepackage{hyperref}       
\usepackage{url}            
\usepackage{booktabs}       
\usepackage{amsfonts}       
\usepackage{nicefrac}       
\usepackage{microtype}      
\usepackage{graphicx}
\usepackage{doi}
\usepackage{amsmath}
\usepackage {vtable}

\newcommand{\reffig}[1]{Figure~\ref{#1}}

\newcommand{\reftab}[1]{Table~\ref{#1}}
\newcommand{\refeq}[1]{\eqref{#1}}
\newcommand{\bi}[1]{\ensuremath{\boldsymbol{#1}}}

\title{Super-resolving 2D stress tensor field conserving equilibrium constraints using physics informed U-Net}

\author  { \href{https://orcid.org/0000-0002-1955-069X}{\includegraphics[scale=0.06]{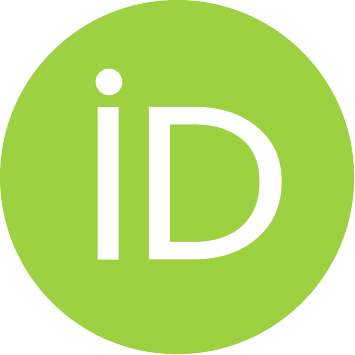}\hspace{1mm}Kazuo Yonekura} \\
	Department of Systems Innovation\\
	The University of Tokyo\\
	Tokyo, JAPAN 113-8656 \\
	\texttt{yonekura@struct.t.u-tokyo.ac.jp} \\
	\AND
	Kento Maruoka \\
	Department of Systems Innovation\\
	The University of Tokyo\\
	Tokyo, JAPAN 113-8656 \\
	\AND
	Kyoku Tyou \\
	Department of Systems Innovation\\
	The University of Tokyo\\
	Tokyo, JAPAN 113-8656 \\
	\AND
	Katsuyuki Suzuki \\
	Department of Systems Innovation\\
	The University of Tokyo\\
	Tokyo, JAPAN 113-8656 \\
}

\date{}

\renewcommand{\shorttitle}{\textit{arXiv} Template}

\hypersetup{
	pdftitle={A template for the arxiv style},
	pdfsubject={q-bio.NC, q-bio.QM},
	pdfauthor={David S.~Hippocampus, Elias D.~Striatum},
	pdfkeywords={First keyword, Second keyword, More},
}

\begin{document}
	\maketitle
	
	\begin{abstract}
In a finite element analysis, using a large number of grids is important to obtain accurate results, but is a resource-consuming task. Aiming to real-time simulation and optimization, it is desired to obtain fine grid analysis results within a limited resource. This paper proposes a super-resolution method that predicts a stress tensor field in a high-resolution from low-resolution contour plots by utilizing a U-Net-based neural network which is called PI-UNet. In addition, the proposed model minimizes the residual of the equilibrium constraints so that it outputs a physically reasonable solution. The proposed network is trained with FEM results of simple shapes, and is validated with a complicated realistic shape to evaluate generalization capability. Although ESRGAN is a standard model for image super-resolution, the proposed U-Net based model outperforms ESRGAN model in the stress tensor prediction task.
	\end{abstract}


\section{Introduction }
With huge advances in computational mechanics, finite element analysis is widely used for designing and optimizing structures. 
Recently, to  obtain accurate computation results, extensive grids are being  used for finite element method (FEM) analysis \cite{Lei19,Nakazawa19,Yadav14} as well as structural optimization \cite{Aage17,Mukherjee21} because the accuracy of FEM is better with fine mesh than with coarse mesh. 
However, the larger the FEM model becomes, the more computation time and memory are required. 

Reduction of computation time in FEM is desired from the points  of view of several applications. 
Real-time FEM simulation has recently become available owing to advances in computation resources and software \cite{Marinkovic19}. Additionally, FEM-based real-time simulation software is commercially available \cite{Fleischmann19}. 
However, such applications do not allow an extensive number of grids because only a little computation time is allowed.  
Hence, the accuracy of the FEM analysis is relatively low. 
Therefore, it is desirable to obtain accurate results with coarse mesh .

Short computation time is also desired in shape and topology optimization. 
In this regard, the FEM analysis is iteratively conducted to calculate the sensitivity from the FEM results. FEM computation time directly affects the total computation time; hence, short computation time is desired. 
Recently, machine learning models have been utilized to realize a shorter FEM computation time ; \cite{Tan21} proposed a surrogate model-based method to approximate the FEM results. 

In computational mechanics research, the superconvergent patch recovery (SPR) method \cite{KUMAR17,Zienkiewicz92a} is proposed to obtain accurate results from coarse mesh.  The SPR method uses a shape function to interpolate stress fields. The accuracy of the SPR method is guaranteed, but the computation time is not small because it needs matrix products of shape functions. 

Additionally, a deep neural network (DNN) model is also utilized to realize the aforementioned aim. 
The advantage of DNN is that the computation time for inference is  short, e.g. less than a few seconds. However, the accuracy of the result is not guaranteed. 
One approach using DNN  directly predicts contour plots from boundary conditions. 
\cite{Jiang21} proposed a StressGAN that predicts two-dimensional  stress contour. 
Another approach is to conduct a super-resolution of the contour plots from coarse FEM results. 
\cite{Tan21} proposed a convolution neural network model to predict fine-mesh results from coarse-mesh results. However, the training and test datasets are similar and generalization ability is not well studied. 
In those existing literature models output only scalar fields, such as von Mises stress or strain fields. 
However, in FEM analysis, it is required to predict stress tensor components. 
In addition, the results should satisfy the physical equations, which is not considered in the existing models do not satisfy them. In particular, the generated contour image does not satisfy equilibrium equations. 
In this study, a DNN-based super-resolution model for a two-dimensional stress tensor field considering equilibrium constraints is proposed. 

Usually, in a super-resolution task, the pixel resolution is improved. However, in the FEM super-resolution task, the pixel resolution need not need be improved. 
The contour image is shown on a FEM mesh, and the resolution of the contour should be improved. 

In image processing research, several super-resolution models have been proposed. 
The most successful models are SRGAN \cite{Ledig17} and ESRGAN \cite{Wang18}, which are based on the generative adversarial network (GAN) \cite{Goodfellow14} framework. 
One of the issues in SRGAN and ESRGAN is that training the models is generally unstable \cite{RankSRGAN}. 
Therefore, in this study, a non-GAN-based model is utilized to avoid training instability.  
The U-Net \cite{UNet} is a fully convolution network model, and U-Net++ \cite{UNet++} is a modified model of U-Net that outperforms U-Net.
The U-Net was first proposed for semantic segmentation of medical images. However, in this study, it is utilized for the super-resolution task. 
The U-Net-based models are non-GAN-based and can be trained more stably than GAN-based models.
In the semantic segmentation task, each pixel is classified for its corresponding classes, e.g., sky, road, or human. Hence, the output is a binary value on each pixel. 
By modifying the U-Net, the model outputs continuous values corresponding to stress tensor components. Thus, it the model can be utilized for super-resolution of stress tensor components. 
In this study, U-Net based models are proposed and compared with ESRGAN-based models. 

The super-resolved images should satisfy the equilibrium equations. 
To obtain such results, the physical constraints are considered using the framework of a physics-informed  neural network (PINN) \cite{PINN,Willard20}. 
One of the methods to consider physical equations is to add the residual of the physical equation to a loss function. For example, \cite{Zhang20} used the residual of Newton's second law to predict seismic response, and \cite{Jia21} used the residual of energy conservation to predict lake temperature. 
The stress tensor follows equilibrium equations. In this study, the residual of the equilibrium equations is employed as the physical loss function. By employing the physical loss, the obtained super-resolved image satisfies the equilibrium constraint. 

This study is organized as follows. The overview of the proposed model and dataset is explained in section 2. 
The proposed physics-informed models are introduced in section 3. 
Numerical experiments are presented in section 4. 
Finally, conclusions follow in section 5.

\section{Proposed model overview and dataset }
\subsection{Problem definition and model overview}
Herein, the task is to obtain reasonable smooth contour images from coarse FEM results. 
The super-resolution model inputs stress tensor data from a coarse mesh and outputs super-resolved  data. 
In this study, only two-dimensional images were used. 
The stress tensor $\sigma$ in the two-dimensional analysis is written as 
\begin{align}
	\sigma = 
	\begin{pmatrix}
		\sigma_x & \tau_{xy} \\
		\tau_{xy}  & \sigma_y
	\end{pmatrix}.
\end{align}

\begin{figure}[h]
	\begin{center}
		\begin{minipage}[h]{\textwidth}
			\begin{center}
				\includegraphics[width=\linewidth]{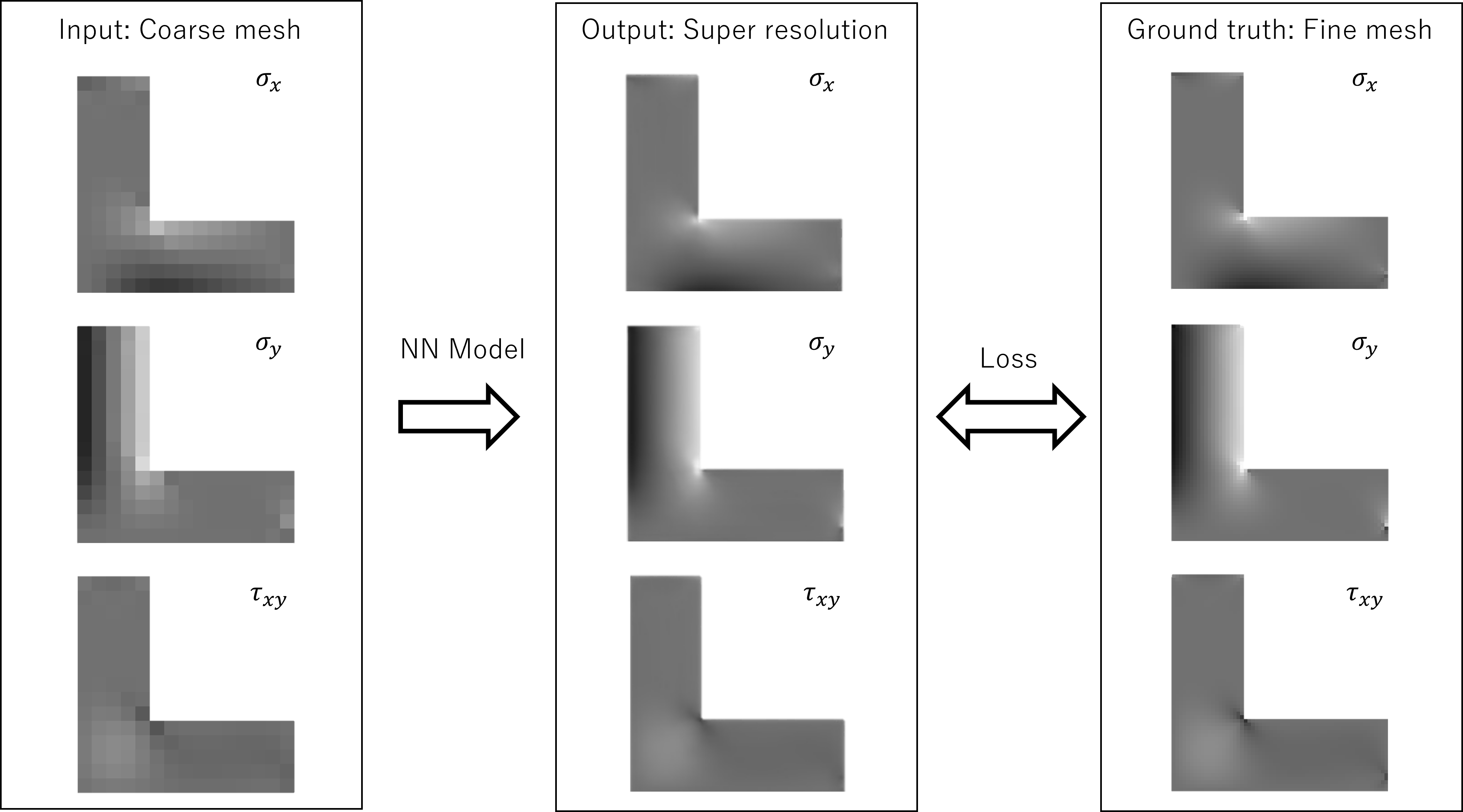}
			\end{center}
		\end{minipage}%
		\caption{Overview of the proposed model.}
		\label{fig:overview}
	\end{center}
\end{figure}

An overview of the proposed method is shown in \reffig{fig:overview}. 
The neural network (NN) model inputs coarse mesh images of three stress tensor components and outputs super-resolved images. 
The loss function is calculated between super-resolved images and ground truth images.
To handle non-uniform shapes, such as rectangles and L-shapes, data are given as pixel images. 
The input coarse images and ground truth fine images are obtained using the FEM analysis. 
It is noted that even in the fine mesh result images, the mesh size is generally larger than the pixel size. 
It is desirable if the mesh resolution of the output super-resolved images is finer than that of ground truth.

\subsection{Dataset  preparation}
FEM analysis was conducted using commercial FEM software, and gray-scale contour images are obtained. 
The contour range was defined when generating the contour images. 
The intensity of pixels was calculated by the post-processing  module based on the stress components and contour map.
We used the linear contour map to calculate stress components from the intensity of images; 
the stress components were calculated by multiplying a coefficient $C_k$ with intensity $I^{1,2,3}_k(i,j)$ at each pixel $(i,j)$;
\begin{align}
	&	\sigma_x (i,j)|_k = C_k \left( I^1_k(i,j) + s_k \right), \\
	&	\sigma_y (i,j)|_k = C_k \left( I^2_k(i,j) + s_k  \right), \\
	&	\tau_{xy} (i,j)|_k = C_k \left( I^3_k(i,j) + s_k  \right),
\end{align}
where $i$ and $j$ are the locations of the pixel; $k$ is the suffix of the image; and $s_k$ is a constant for adjusting contour level. 
The superscripts $1$, $2$, and $3$ on $I$ correspond to $\sigma_x$, $\sigma_y$, and $\tau_{xy}$, respectively. 
Consequently, $I^1_k(i,j)$ is the intensity at the pixel $(i,j)$ of $k$-th image corresponding to $\sigma_x$. 
$C_k$ and $s_k$ are defined for each analysis case $k$, which implies that the contour maps are the same for $\sigma_x$, $\sigma_y$, and $\tau_{xy}$ of one analysis, but different between different analysis cases. 
The data consists of a set of coarse mesh results and fine mesh results, i.e., three coarse contour images and three fine contour images.

\begin{figure}[h]
	\begin{center}
		\begin{minipage}[h]{0.5\textwidth}
			\begin{center}
				\includegraphics[width=\linewidth]{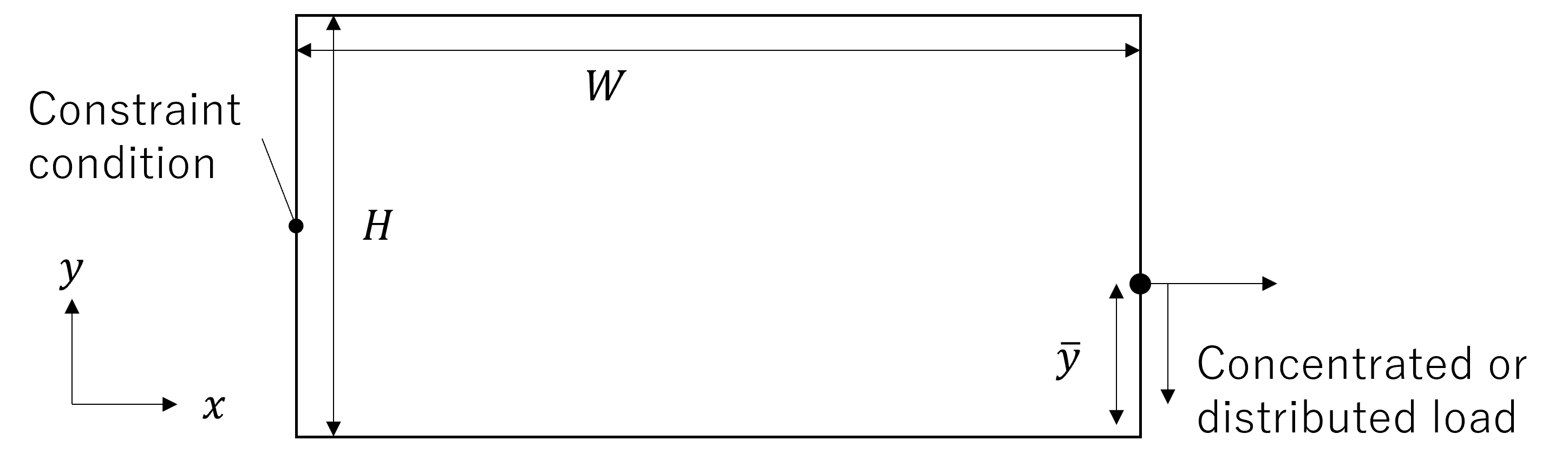}
				\par {(a) Cantilever data.}
			\end{center}
		\end{minipage}%
		\begin{minipage}[h]{0.5\textwidth}
			\begin{center}
				\includegraphics[width=\linewidth]{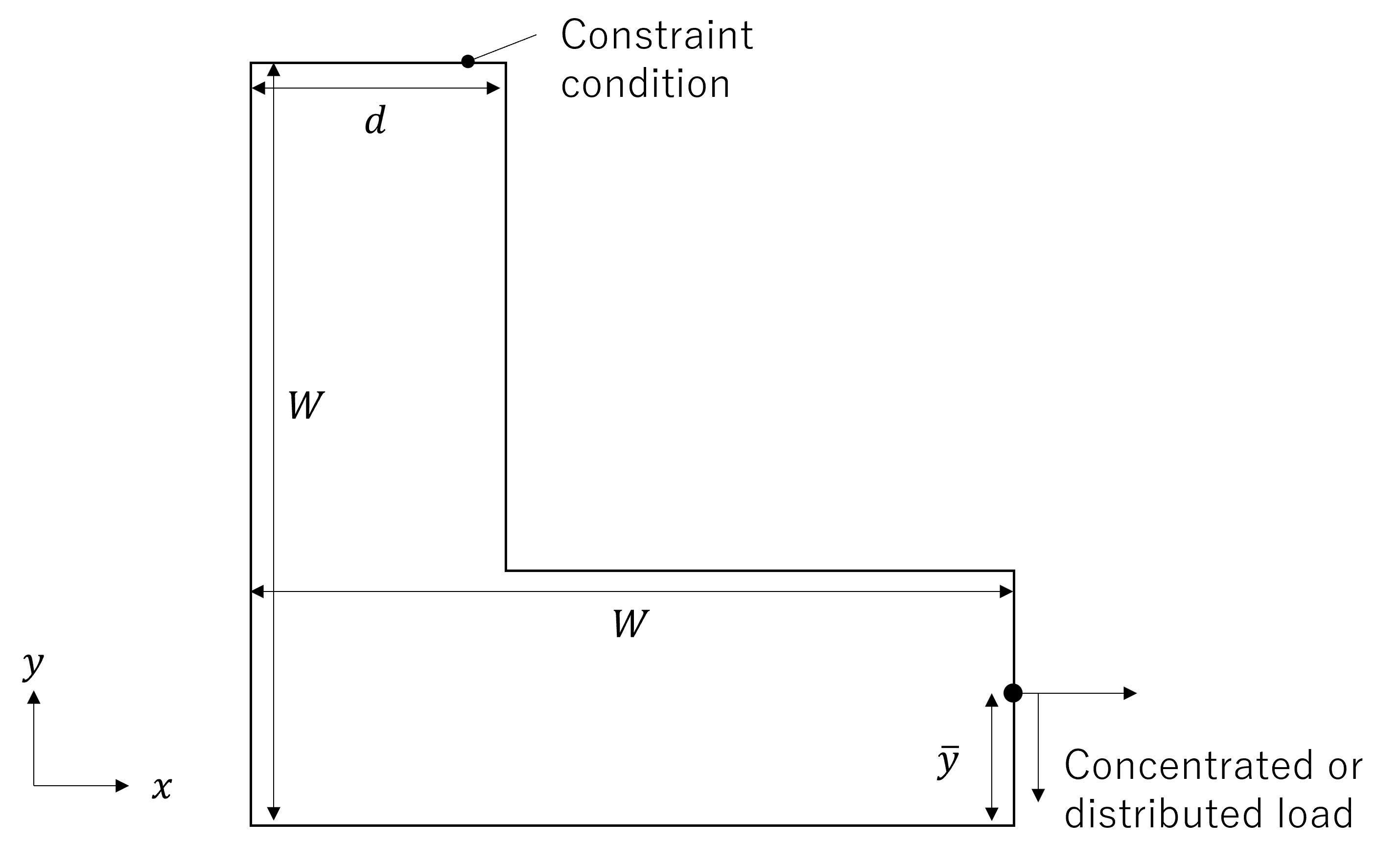}
				\par  {(b) L-shape data.}
			\end{center}
		\end{minipage}%
		\par
		\begin{minipage}[h]{0.5\textwidth}
			\begin{center}
				\includegraphics[width=\linewidth]{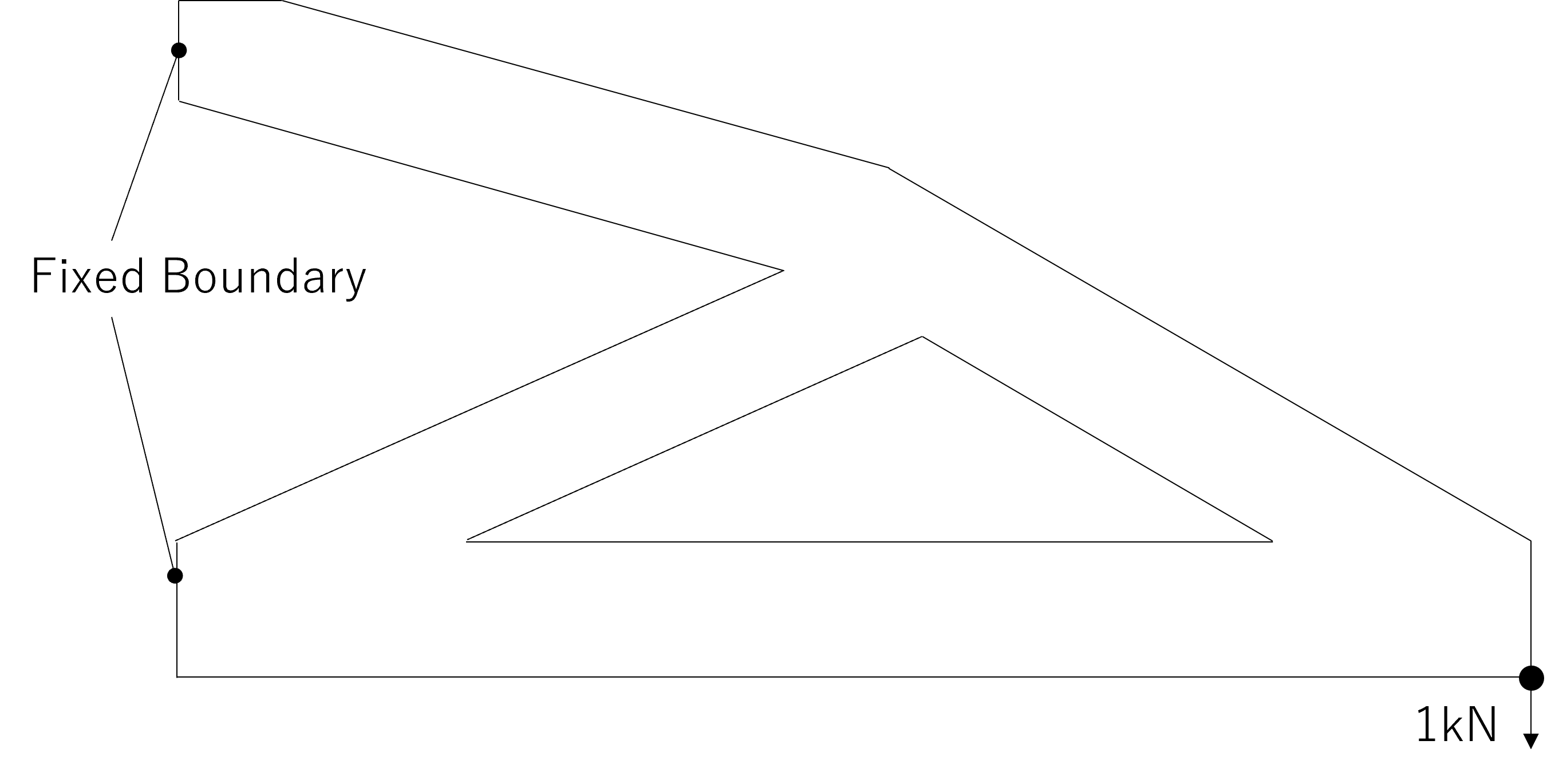}
				\par	{(c) Truss-like cantilever.}
			\end{center}
		\end{minipage}%
		\caption{Close up view of coarse, fine, and super-resolved images.}
		\label{fig:data}
	\end{center}
\end{figure}

The dataset comprises training and validation datasets. 
From an application point of view, it is important that a model trained with a small number of data is valid for a wide range of analyses. 
Hence, in this study, we trained the model using relatively simple data and validated it with a complicated truss-like structure. 
The training dataset was composed of simple analysis cases, i.e., cantilevers and L-shaped beams, 
whereas the validation data was a truss-like cantilever; both datasets are shown in \reffig{fig:data}. 
Linear quadrilateral element was used. The mesh size $\ell$ was $\ell = H/10$ and $ \ell= d/5$ in the coarse mesh analysis, and $\ell = H/40$ and $ \ell= d/20$ in the fine mesh analysis; essentially, the fine mesh was $4$ times smaller than that of the coarse mesh.
Data were generated using the same shape on different boundary conditions and loads. 
In the cantilever and L-shape, the constraint condition was selected from two types, i.e., fixed boundary and sliding boundary. 
The loads were selected from multiple cases listed in \reftab{tab:model}. The magnitude of total loads was $1kN$ in all cases. 
The dimensions were set as $W=2H$ and $L=3d$. 
After preparing the FEM results, data augmentation was performed to increase the number of data. 
The image was inverted vertically and horizontally. 
The contour image of the inside shape, i.e., only the contour area excluding the white background area, was also inverted. Essentially, black was inverted into white. 
The total number of datasets for cantilever and L-shape was $504$, including augmented data. 
The dataset was randomly split into $3:1$ for training and test data. 
A truss-like cantilever is employed as the validation dataset, which has a similar shape to the topology-optimized shape, to determine if the model could be applied to topology optimization. 
The element type in the validation dataset was triangle to determine if the model could work with different types of elements.
For data augmentation, left-right, top-bottom,  and black-white flips were enabled. 

\begin{table}[htpb]
	\begin{center}
		\caption{Loading conditions of FEM models}
		\label{tab:model}       
		\begin{minipage}[h]{0.4\textwidth}
			\begin{center}
				{\footnotesize 
					Cantilever\\
					\begin{tabular}{r|r|c}
						\hline
						Load type & Direction & $\bar{y}$ \\
						\hline
						Concentrated & $x$ & $id/5$ \\ 
						Concentrated & $y$ & $id/5$ \\ 
						Distributed  & $x$ & --- \\ 
						Distributed  & $y$ & --- \\
						\hline
					\end{tabular}%
					\\
					\qquad \qquad \qquad \qquad ($i=0,1,\dots,5$)
				}
			\end{center}
		\end{minipage}
		\begin{minipage}[h]{0.5\textwidth}
			\begin{center}
				{\footnotesize 
					L-shape \\
					\begin{tabular}{r|r|c}
						\hline
						Load type & Direction & $\bar{y}$ \\
						\hline
						Concentrated & $x$ & $iH/10$ \\ 
						Concentrated & $y$ & $iH/10$ \\ 
						Distributed  & $x$ & --- \\ 
						Distributed  & $y$ & --- \\
						\hline
					\end{tabular}%
					\\
					\qquad \qquad \qquad \qquad \qquad ($i=0,1,\dots,5$)
				}
			\end{center}
		\end{minipage}
	\end{center}%
\end{table}%

\section{Physics informed super-resolution models } \label{model}
\subsection{Physical loss}
The stress tensor $\sigma$ has to satisfy the equilibrium constraint;
\begin{align}
	&{\rm div} \sigma + \boldsymbol{f} = 
	\begin{pmatrix}
		\frac{\partial \sigma_x}{ \partial x} + \frac{\partial \tau_{xy} }{ \partial y} + f_x\\
		\frac{\partial \sigma_{y}}{ \partial y} + \frac{\partial \tau_{xy} }{ \partial x} + f_y
	\end{pmatrix}
	= \boldsymbol{0},  \label{eq.div}
\end{align}
where $\boldsymbol{f} = (f_x, f_y)^{\top}$ implies a body force. 
The physics loss $\mathcal{L}_{\rm PL}$ is defined as a penalty for violating equation \eqref{eq.div}.
\begin{align}
	\mathcal{L}_{\rm PL} = \| {\rm div} \sigma + \boldsymbol{f} \|^2 . \label{eq.L_PG}
\end{align}

To calculate derivatives in equation \ref{eq.div} from output data $\boldsymbol{y} = (y_1, y_2, y_3)^\top$, the central difference is applied on pixels. 
The derivatives at pixel $(i,j)$ are calculated by
\begin{align}
	&	\frac{\partial \sigma_x}{ \partial x}(i,j) = y_1(i+1,j) - y_1(i-1,j), \\
	&	\frac{\partial \sigma_y}{ \partial y}(i,j) = y_2(i,j+1) - y_2(i,j-1), \\
	&	\frac{\partial \tau_{xy}}{ \partial x}(i,j) = y_3(i+1,j) - y_3(i-1,j), \\
	&	\frac{\partial \tau_{xy}}{ \partial y}(i,j) = y_3(i,j+1) - y_3(i,j-1).  
\end{align}

The loss $\mathcal{L}_{\rm PL}$ must be summed only the internal region of the shapes. 
Hence, the mask function is defined to extract the internal region.
\begin{align}
	m(i,j) = 1 {\rm ~if~} y_1(i+1) > 1 - \varepsilon
\end{align}
and $ \hat{\mathcal{L}}_{\rm PL} =  \sum_{i,j} m(i,j) \mathcal{L}_{\rm PL} (i,j)$. 
In the dataset, the external loads $\boldsymbol{f}$ are applied only on the boundary edges and points. Those edges and points are masked by the mask function. Hence, $\boldsymbol{f}$ in \refeq{eq.L_PG} can be eliminated.

It is obvious that $\hat{\mathcal{L}}_{\rm PL}$ is minimum when $\boldsymbol{y}$ is constant everywhere. However, in such a case, the MSE loss becomes large. 

\subsection{Modified U-Net/U-Net++ and PI-UNet/PI-UNet++}
\begin{figure}[h]
	\begin{center}
		\begin{minipage}[h]{0.5\textwidth}
			\begin{center}
				\includegraphics[width=\linewidth]{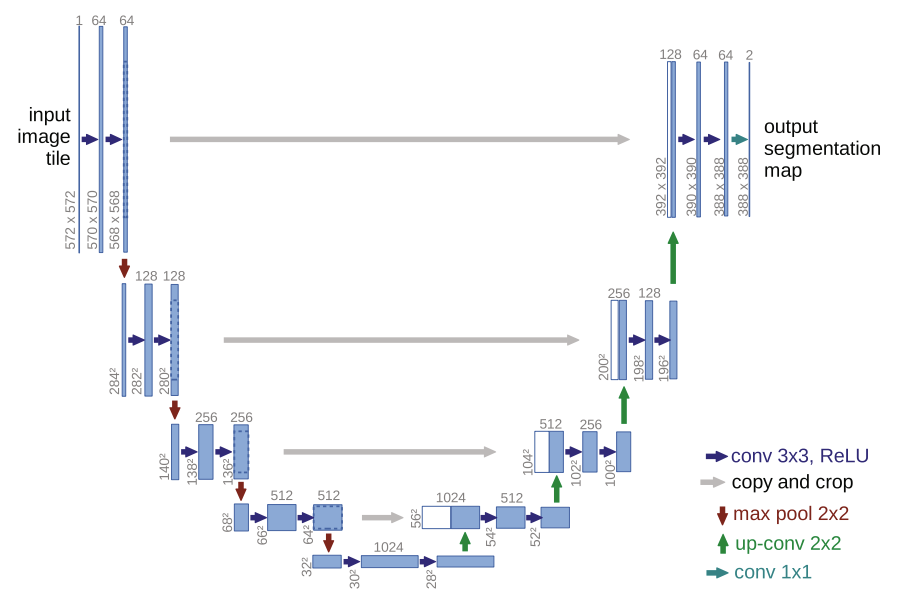}
				\par {(a) U-Net architecture. (Cited from \cite{UNet} Fig.1)}
			\end{center}
		\end{minipage}%
		\begin{minipage}[h]{0.5\textwidth}
			\begin{center}
				\includegraphics[width=\linewidth]{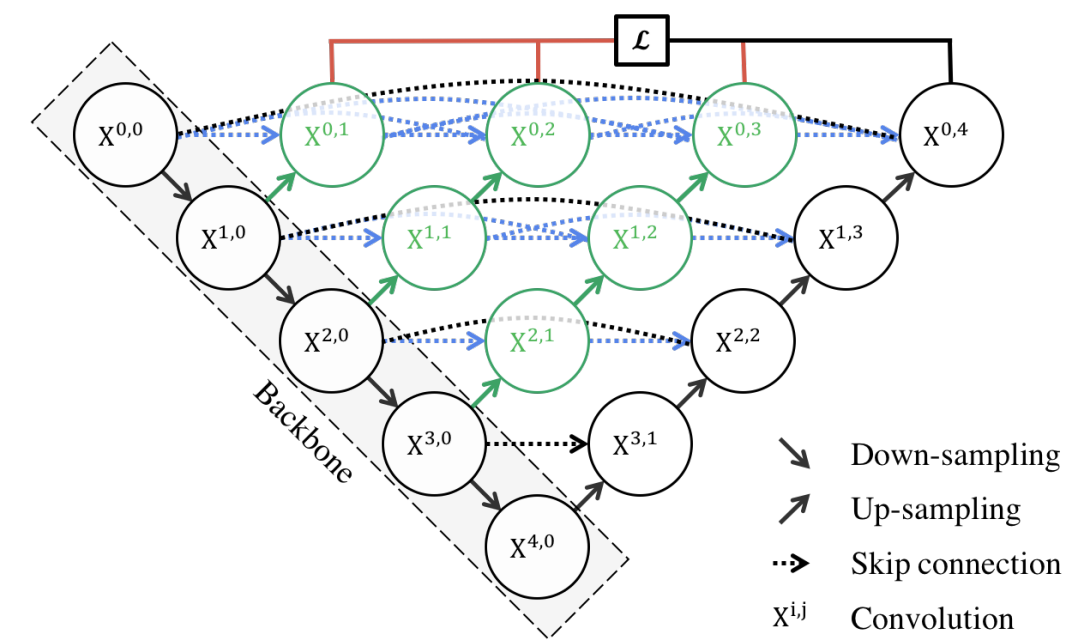}
				\par {(b) U-Net++ architecture. (Cited from \cite{UNet++} Fig.1 (a))}
			\end{center}
		\end{minipage}%
		\caption{Architecture of U-Net and U-Net++.}
		\label{fig:UNet_arch}
	\end{center}
\end{figure}

U-Net \cite{UNet} is an encoder-decoder model that is proposed for semantic segmentation for medical images, which is not usually used for super-resolution. 
In this study, the U-Net is slightly modified to allow its applicability for super-resolution purpose. 
The architecture of U-Net is shown in \reffig{fig:UNet_arch}(a). 
The input data is encoded to a low-level feature map using the max-pooling  layer and the feature map is decoded to the original resolution. 
The feature map of each encoder level is connected to the feature map of the same decoder level. 

The U-Net NN inputs images and outputs the segmentation map. The segmentation map consists of binary data on each pixel, i.e., true/false with respect to the corresponding classes. 
The softmax cross-entropy  is employed as the loss function in the U-Net. 

In this study, the aim is to predict the intensity of each pixel whose value is in $[0,1]$ from original images. 
Hence, we removed the softmax function and used the output from the sigmoid function at the last layer so that the model outputs the value in $[0,1]$. 
In addition, the loss function is modified and the mean squared error (MSE) is used;
\begin{align}
	\mathcal{L}_{\rm MSE} = \frac{1}{ N } \left\| \bi{t} - \bi{o} \right\|^2, 
\end{align}
where $\bi{t}$ and $\bi{o}$ are the ground truth and output data, respectively; and $N$ is the number of pixels. 
In this study, the modified model is called a modified U-Net. 
It is noted that the resolution of both the input and output images is the same, which is a common situation in U-Net. 

In the physics-informed U-Net (PI-UNet), 
the architecture of the model is the same as the modified U-Net, except that physical loss $ \hat{\mathcal{L}}_{\rm PL}$ is added to the loss function of the modified U-Net.
\begin{align*}
	\mathcal{L}_{\rm PI\mathchar`-UNet} =  \mathcal{L}_{\rm MSE} + \hat{\mathcal{L}}_{\rm PL}. 
\end{align*}
The pixel resolution of the output image is the same as that of the input image, which is a normal condition in UNet++. 

UNet++ \cite{UNet++} is a nested U-Net model and is shown in \reffig{fig:UNet_arch}(b). 
The UNet++ adds a decoder from every encoder level to the input level. 
Herein, the U-Net++ outperformed U-Net in the medical image segmentation task.

\subsection{ESRGAN and PI-ESRGAN}
ESRGAN \cite{Wang18} is a GAN-based super-resolution model. 
In the GAN framework, networks consist of a generator network and a discriminator network. The loss function in the GAN is defined as follows.
\begin{align}
	V(D,G) = E_{\bi{x} \sim p_{\rm data} } \left[ \log D( \bi{x} ) \right] + E_{ \bi{z} \sim p_z } \left[ 1 - \log D(G(\bi{z})) \right]
\end{align}
where G and D imply the generator and the discriminator, respectively. $G$ inputs noise $\bi{z}$. 
$\bi{z}$ is a noise vector, $\bi{x}$ is a real data, $p_{\rm data}$ and $p_z$ are the probability distribution of the real data and noise, respectively.
The generator minimizes $V(D,G)$, whereas the discriminator maximizes V. 


In the ESRGAN model, the relativistic discriminator,  proposed in the relativistic GAN \cite{Alexia19}, is employed. 
The ordinal discriminator is described by $D(\bi{x})= {\rm sigmoid} \left( C \left(\bi{x} \right) \right)$, where the function $C$ implies the output from the discriminator without the activation function. Then, the relativistic discriminator, $D_{R_a}$, is defined as
\begin{align}
	D_{R_a}  \left( \bi{x}, \bi{x}' \right) = {\rm sigmoid} \left( C(\bi{x})- E_{\bi{x}'} \left[ C \left( \bi{x}' \right) \right] \right). 
\end{align}
Thus, the loss function of the discriminator network, $L^{\rm D}_{\rm ESRGAN}$, is formulated using the relativistic discriminator as
\begin{align}
	L^{\rm D}_{\rm ESRGAN} = - E_{\bi{x} \sim p_{data}} \left[ \log \left( D_{R_a} \left( \bi{x}, \bi{x}' \right) \right) \right] - E_{\bi{x}' \sim p_{x'}} \left[ \log \left( 1 - D_{R_a}\left(\bi{x}',  \bi{x} \right) \right) \right]. \label{Dra_loss}
\end{align}

The loss function of the generator network in ESRGAN comprises the perceptual loss, GAN loss, and $L_1$ loss;
\begin{align}
	L^{\rm G}_{\rm ESRGAN} = L_{percep} + \lambda L^{R_a}_G + \eta L_1, 
\end{align}
where $\lambda$ and $\eta$ are the hyperparameters to balance each term; 
$\lambda = 0.005$ and $\eta = 0.01$ are used in the numerical examples. 
The perceptual loss is 
\begin{align}
	L_{percep} = \frac{1}{W_{i,j} H_{i,j}} \sum_{x=1}^{W_{i,j}} \sum_{y=1}^{H_{i,j}} \left(  \phi_{i,j} \left( I^{HR} \right)_{x,y} - \phi_{i,j} \left( G \left( I^{LR} \right) \right)_{x,y} \right)^2  , 
\end{align}
where $I^{HR}$ and $I^{LR}$ imply the pixel intensity of the high and low-resolution  image, respectively. 
$\phi_{i,j}$ represents the feature map obtained by the $j$-th convolution before $i$-th max-pooling layer in the pre-trained VGG19 network \cite{VGG19}.  $H_{i,j}$ and $W_{i,j}$ are the dimensions of the corresponding feature maps within the VGG19 network. 
The GAN loss, $L^{R_a}_G$, is formulated as
\begin{align}
	L^{R_a}_G = - E_{\bi{x} \sim p_{data}} \left[ \log \left(1-  D_{R_a} \left( \bi{x}, \bi{x}' \right) \right) \right] - E_{\bi{x}' \sim p_{x'}} \left[ \log \left( D_{R_a}\left(\bi{x}',  \bi{x} \right) \right) \right], 
\end{align}
which has a symmetric form of discriminator loss, $L^{R_a}_D$, in \eqref{Dra_loss}.
The $L_1$ loss is 
\begin{align}
	L_1 = E_{I^{LR}} \| G \left( I^{LR} \right) - \bi{t} \|_1
\end{align}

In the original ESRGAN model, the resolution of the high-resolution  training data is different from that of the low-resolution data. 
However, as previously discussed, both resolutions are set the same in this study. 

In the physics informed ESRGAN (PI-ESRGAN), the loss function of the generator is modified to include physics loss; 
\begin{align*}
	&	\mathcal{L}_{\rm PI\mathchar`-ESRGAN}^{\rm G} =  L^{\rm G}_{\rm ESRGAN} + \hat{\mathcal{L}}_{\rm PL}. 
\end{align*}
The loss function of the discriminator in PI-ESRGAN is the same as in original ESRGAN; $\mathcal{L}_{\rm PI\mathchar`-ESRGAN}^{\rm D} =  L^{\rm D}_{\rm ESRGAN} $.

\section{Numerical experiments }
\subsection{Training and testing on the similar data}
\begin{figure}[h]
	\begin{center}
		\footnotesize
		\begin{tabular}{|C{5cm}{}{m}|C{5cm}{}{t}|C{5cm}{}{t}|C{5cm}{}{t}|}
			\hline
			Model & $\sigma_x$ & $\sigma_y$ & $\tau_{xy}$ \nextRow
			\hline 
			Coarse & \includegraphics[width=0.5\linewidth]{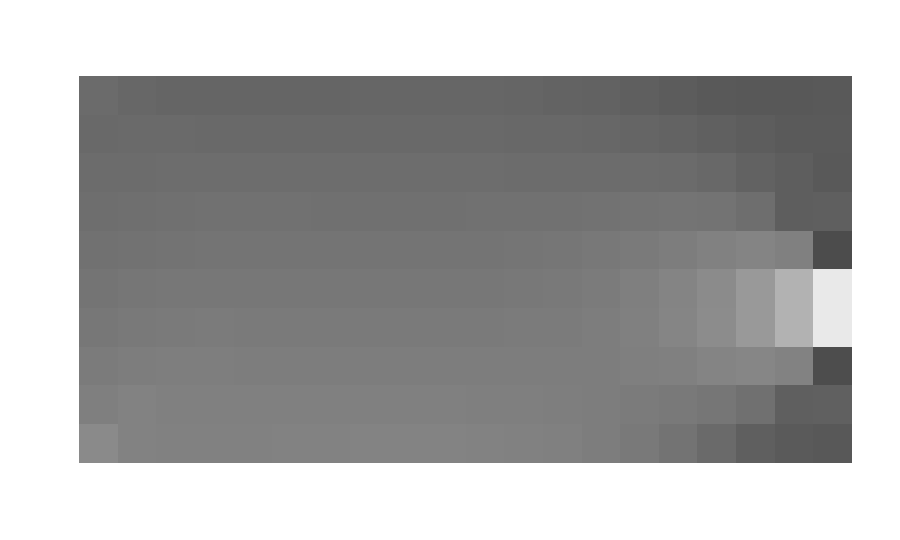} & \includegraphics[width=0.5\linewidth]{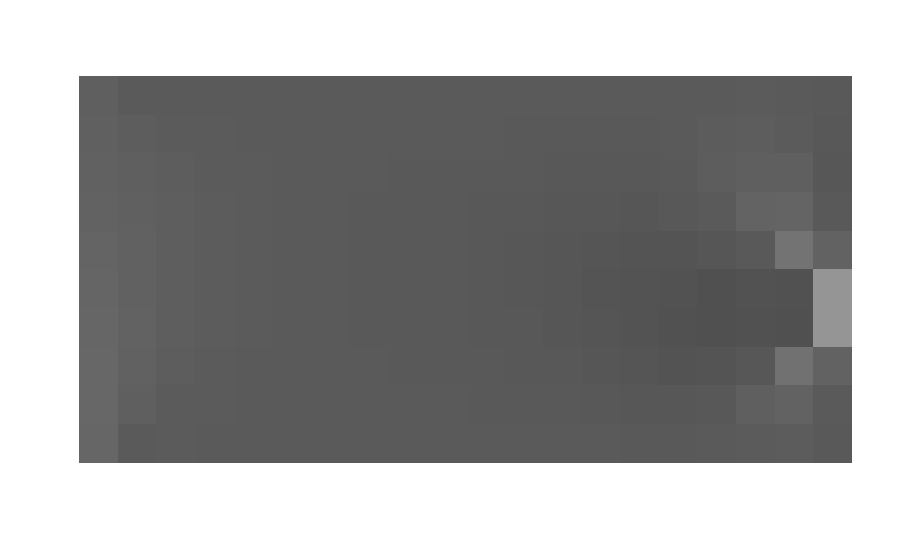} & \includegraphics[width=0.5\linewidth]{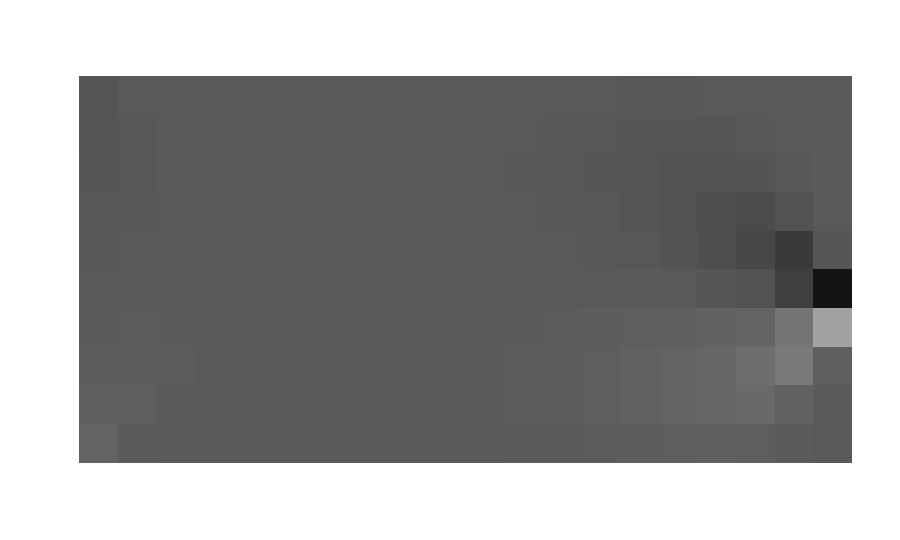}  \nextRow
			\hline 
			ESRGAN & \includegraphics[width=0.5\linewidth]{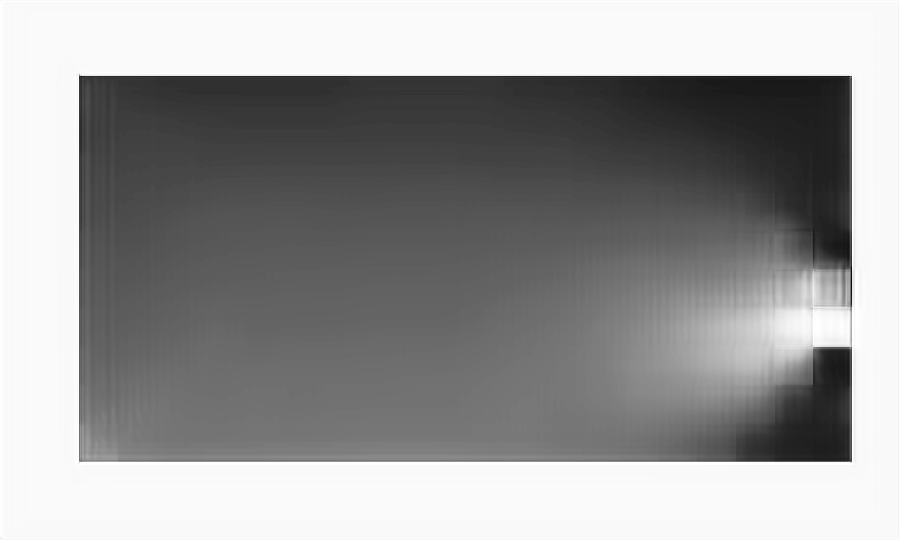} & \includegraphics[width=0.5\linewidth]{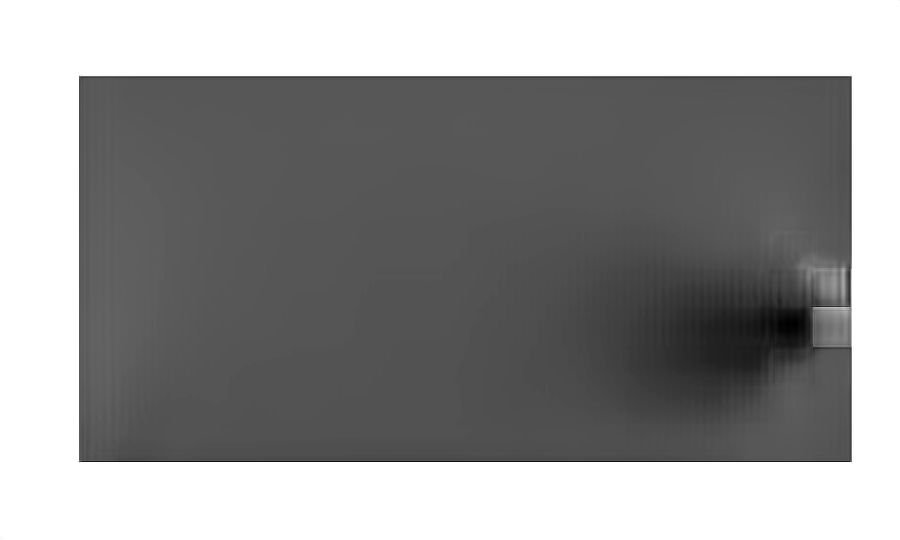} & \includegraphics[width=0.5\linewidth]{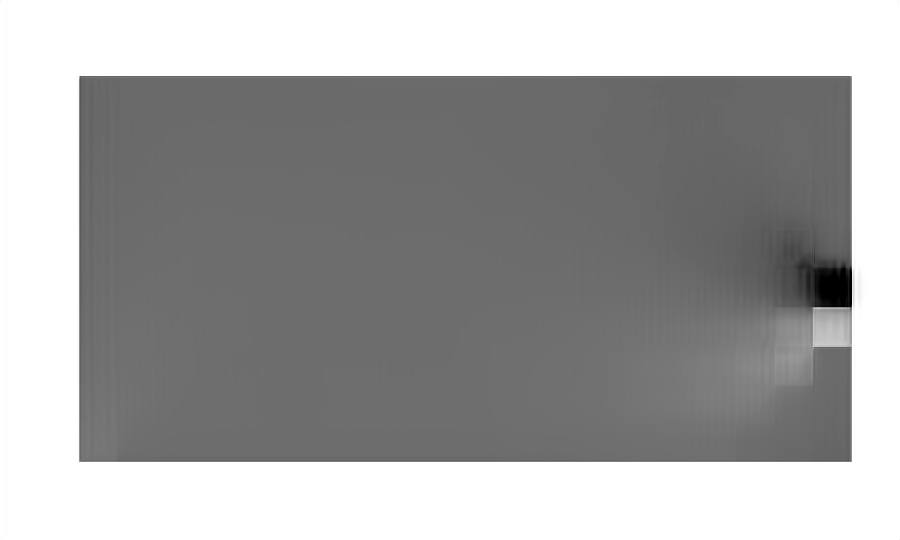}  \nextRow
			\hline
			PI-ESRGAN & \includegraphics[width=0.5\linewidth]{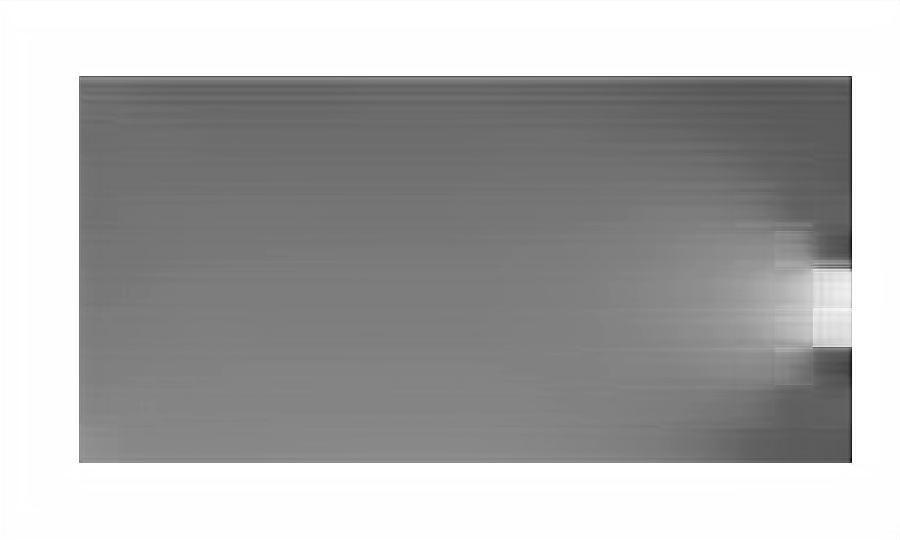} & \includegraphics[width=0.5\linewidth]{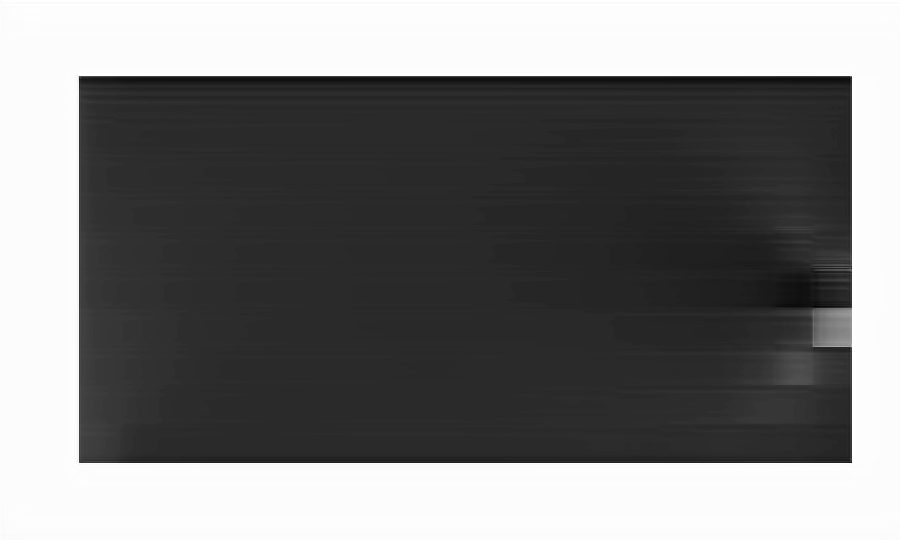} & \includegraphics[width=0.5\linewidth]{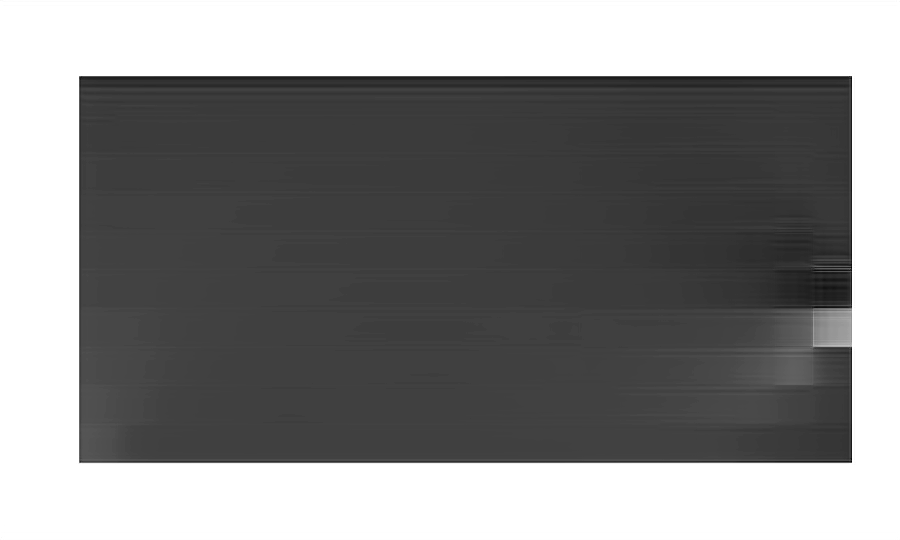}  \nextRow
			\hline
			U-Net & \includegraphics[width=0.5\linewidth]{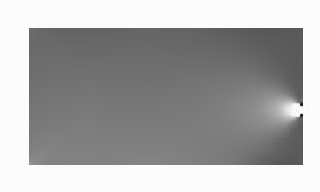} & \includegraphics[width=0.5\linewidth]{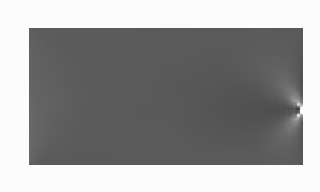} & \includegraphics[width=0.5\linewidth]{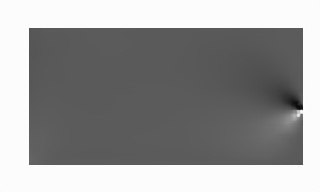}  \nextRow
			\hline
			PI-UNet & \includegraphics[width=0.5\linewidth]{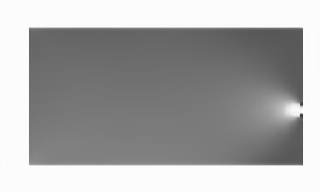} & \includegraphics[width=0.5\linewidth]{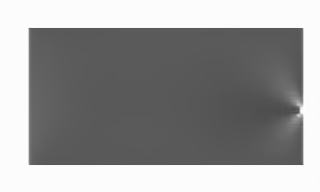} & \includegraphics[width=0.5\linewidth]{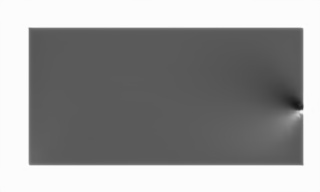}  \nextRow
			\hline
			U-Net++ & \includegraphics[width=0.5\linewidth]{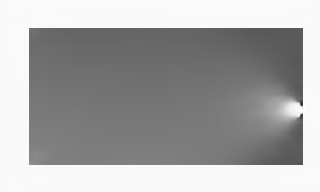} & \includegraphics[width=0.5\linewidth]{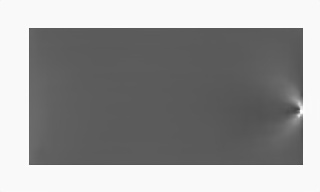} & \includegraphics[width=0.5\linewidth]{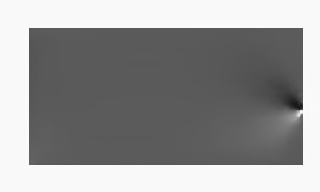}  \nextRow
			\hline
			PI-UNet++ & \includegraphics[width=0.5\linewidth]{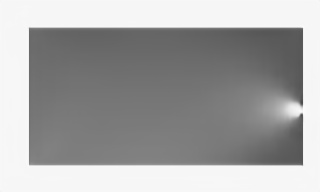} & \includegraphics[width=0.5\linewidth]{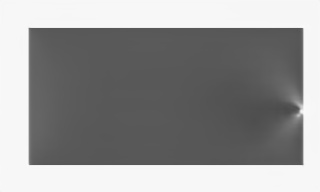} & \includegraphics[width=0.5\linewidth]{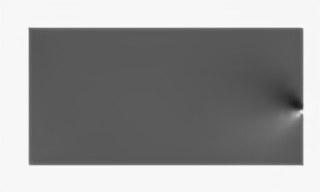}  \nextRow
			\hline
			Fine & \includegraphics[width=0.5\linewidth]{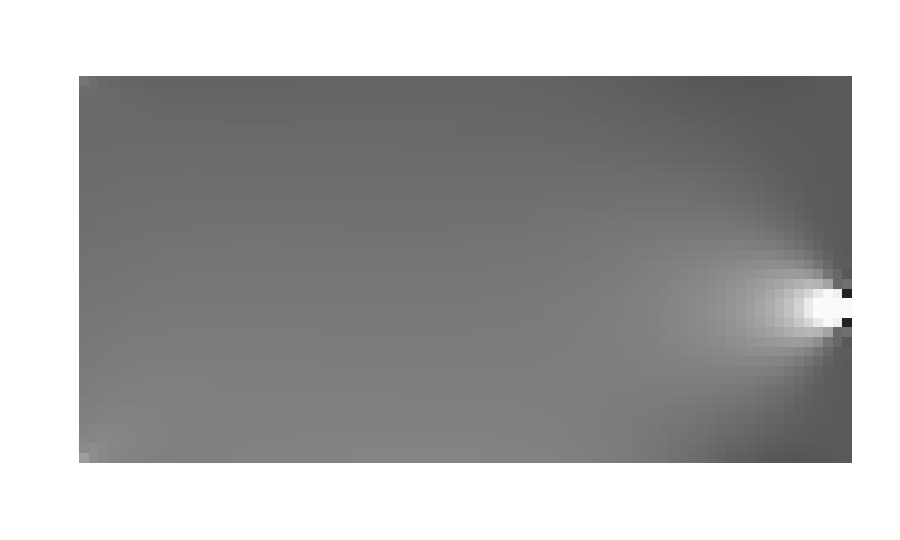} & \includegraphics[width=0.5\linewidth]{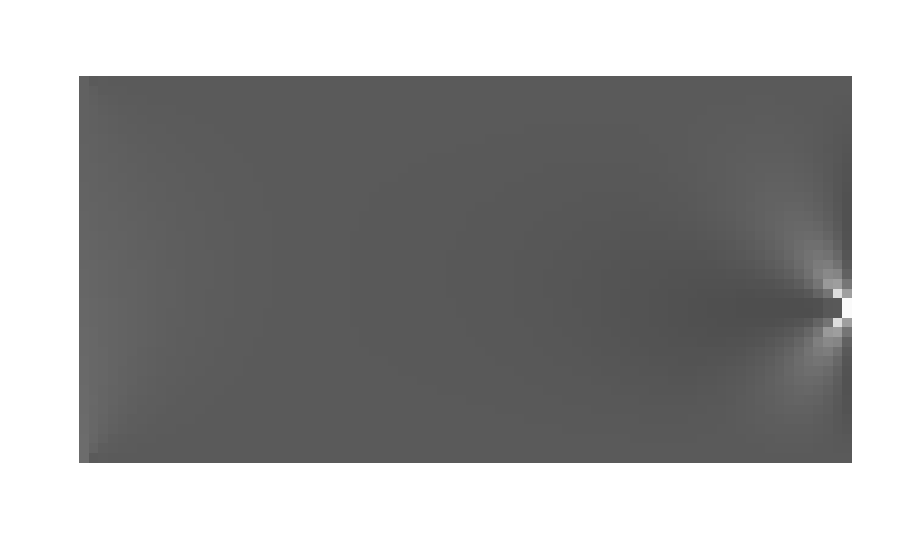} & \includegraphics[width=0.5\linewidth]{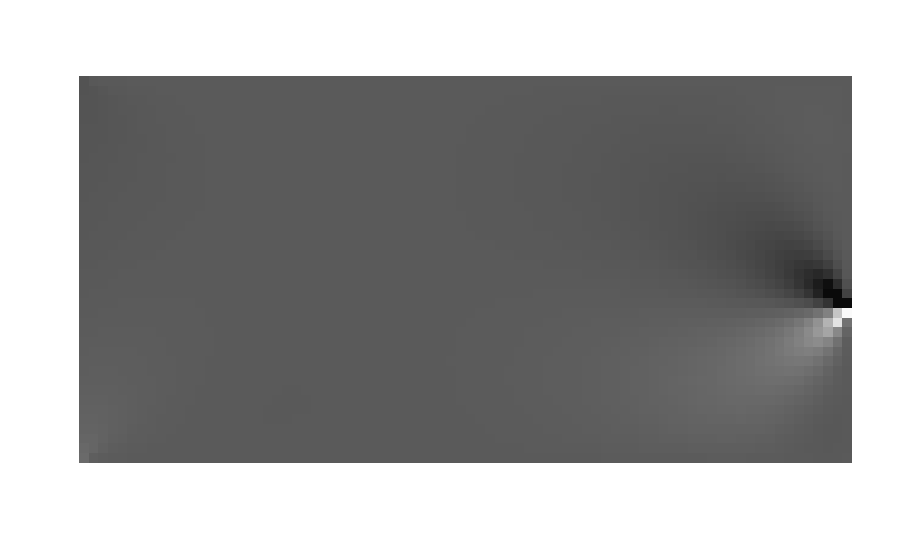}  \nextRow
			\hline
		\end{tabular}
	\end{center}
	\caption{Cantilever example}
	\label{fig:bar_ex}
\end{figure}

\begin{figure}[h]
	\begin{center}
		\footnotesize
		\begin{tabular}{|C{5cm}{}{m}|C{5cm}{}{t}|C{5cm}{}{t}|C{5cm}{}{t}|}
			\hline
			Model & $\sigma_x$ & $\sigma_y$ & $\tau_{xy}$ \nextRow
			\hline 
			Coarse & \includegraphics[width=0.5\linewidth]{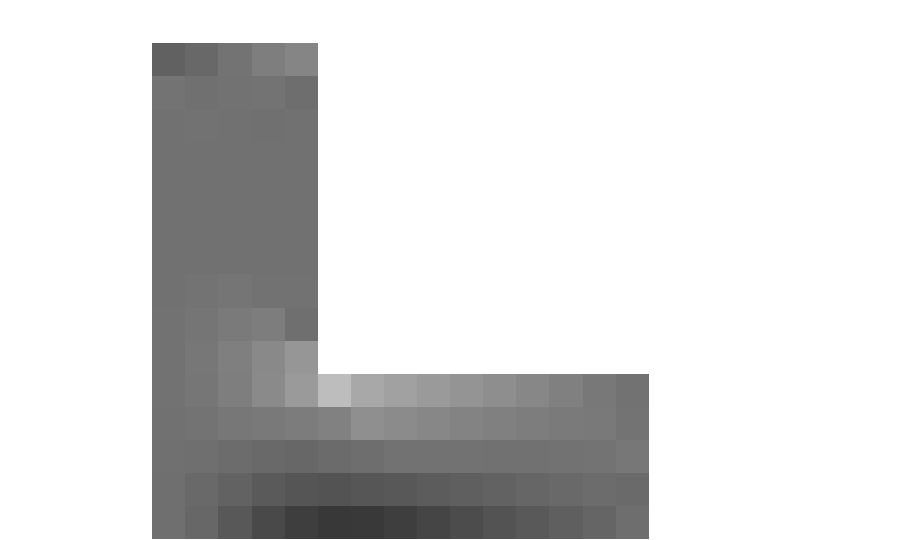} & \includegraphics[width=0.5\linewidth]{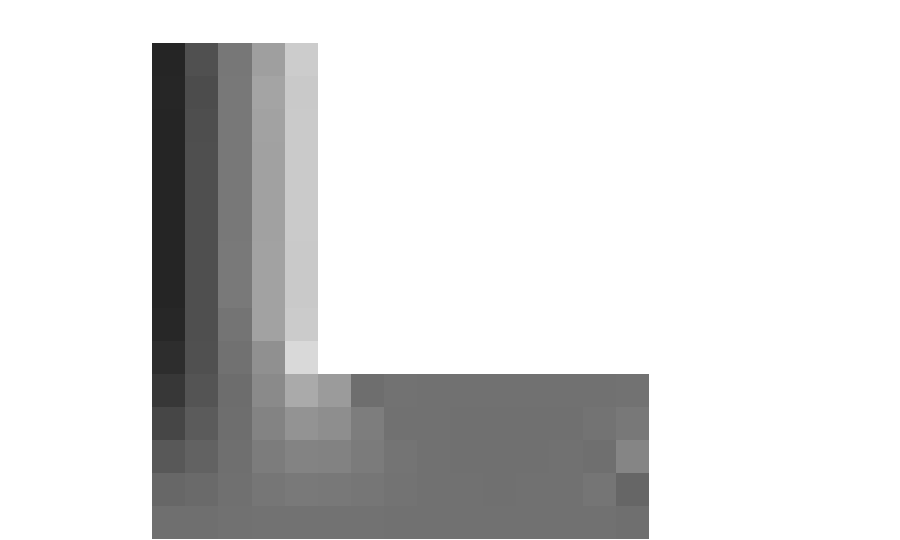} & 
			\includegraphics[width=0.5\linewidth]{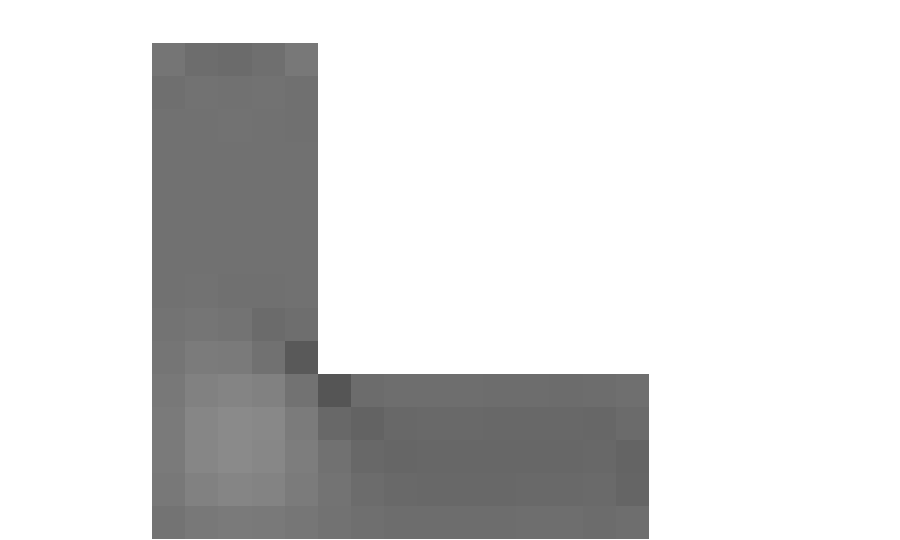}  \nextRow
			\hline 
			ESRGAN & \includegraphics[width=0.5\linewidth]{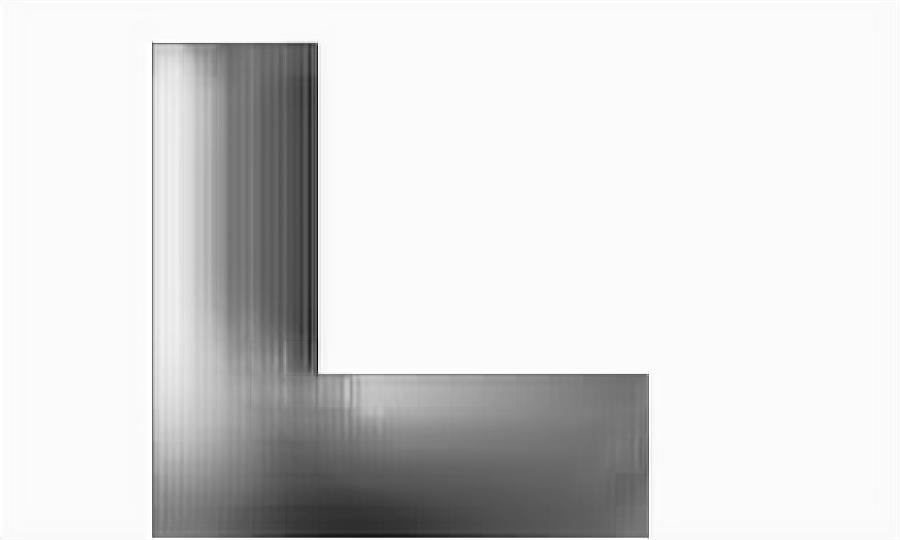} & \includegraphics[width=0.5\linewidth]{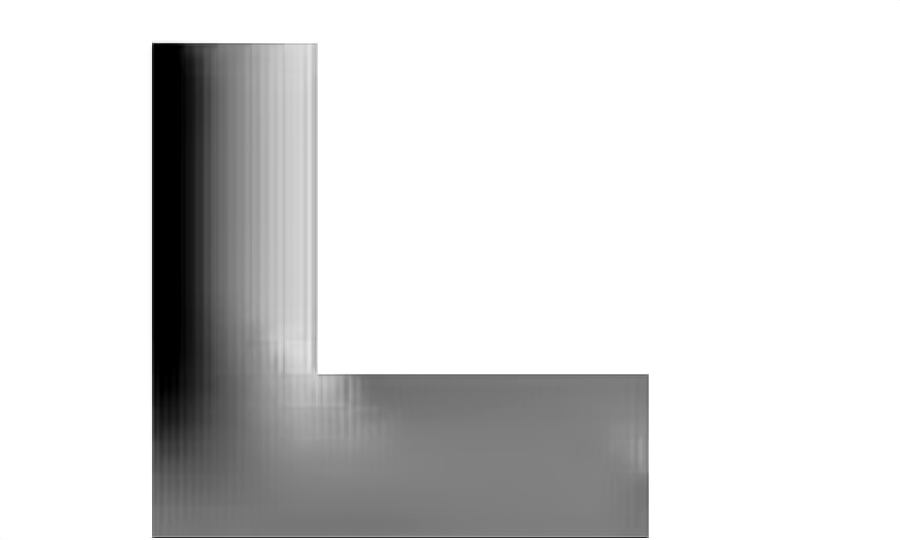} & \includegraphics[width=0.5\linewidth]{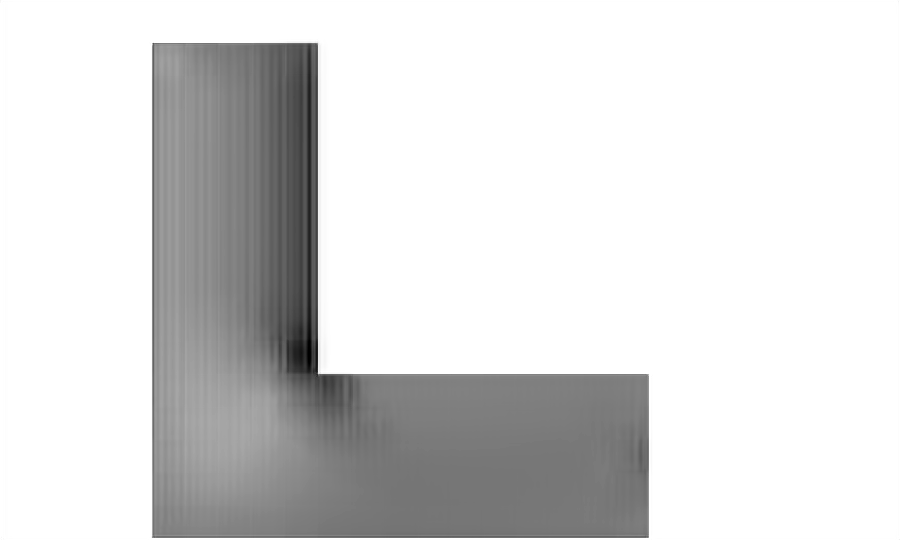}  \nextRow
			\hline
			PI-ESRGAN & \includegraphics[width=0.5\linewidth]{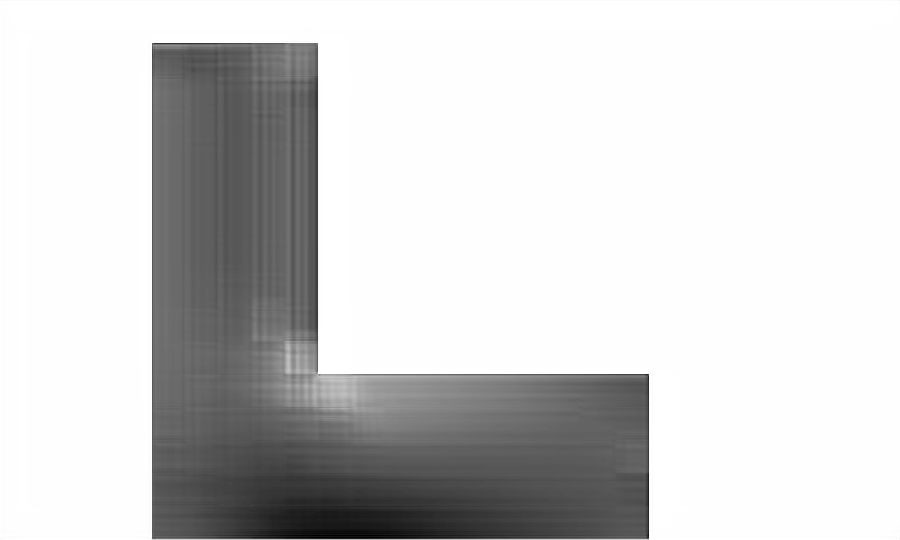} & \includegraphics[width=0.5\linewidth]{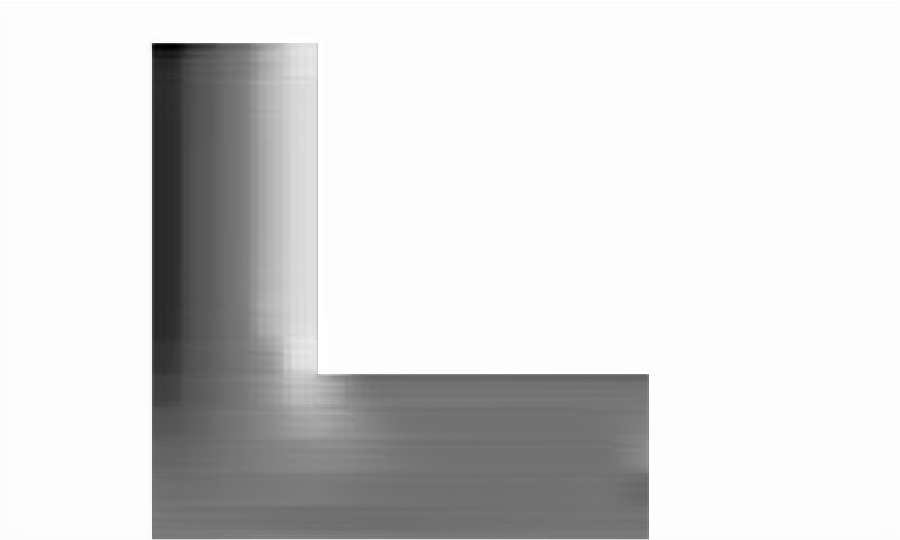} & \includegraphics[width=0.5\linewidth]{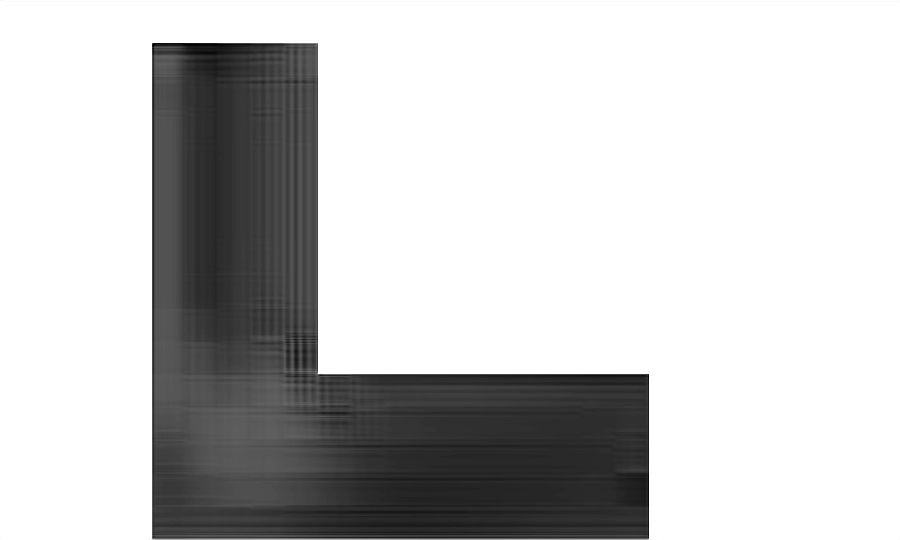}  \nextRow
			\hline
			U-Net & \includegraphics[width=0.5\linewidth]{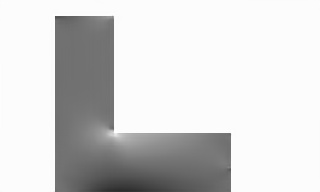} & \includegraphics[width=0.5\linewidth]{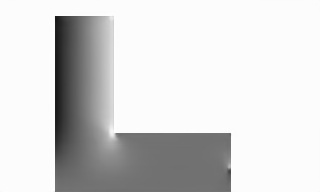} & \includegraphics[width=0.5\linewidth]{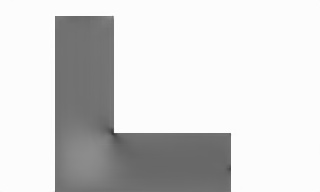}  \nextRow
			\hline
			PI-UNet & \includegraphics[width=0.5\linewidth]{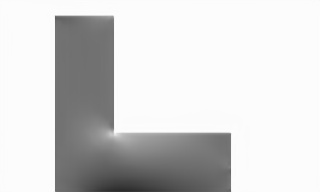} & \includegraphics[width=0.5\linewidth]{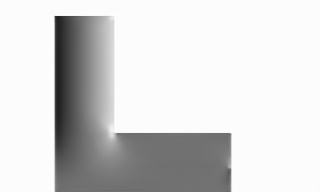} & \includegraphics[width=0.5\linewidth]{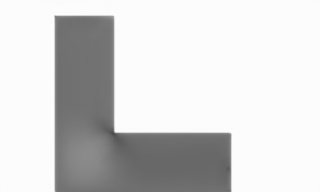}  \nextRow
			\hline
			U-Net++ & \includegraphics[width=0.5\linewidth]{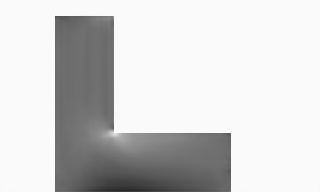} & \includegraphics[width=0.5\linewidth]{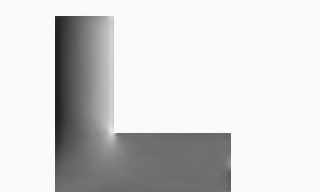} & \includegraphics[width=0.5\linewidth]{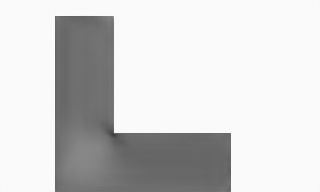}  \nextRow
			\hline
			PI-UNet++ & \includegraphics[width=0.5\linewidth]{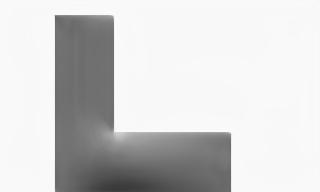} & \includegraphics[width=0.5\linewidth]{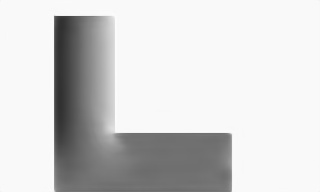} & \includegraphics[width=0.5\linewidth]{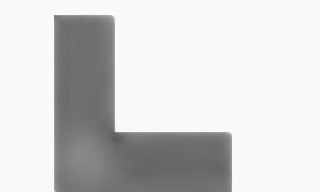}  \nextRow
			\hline
			Fine & \includegraphics[width=0.5\linewidth]{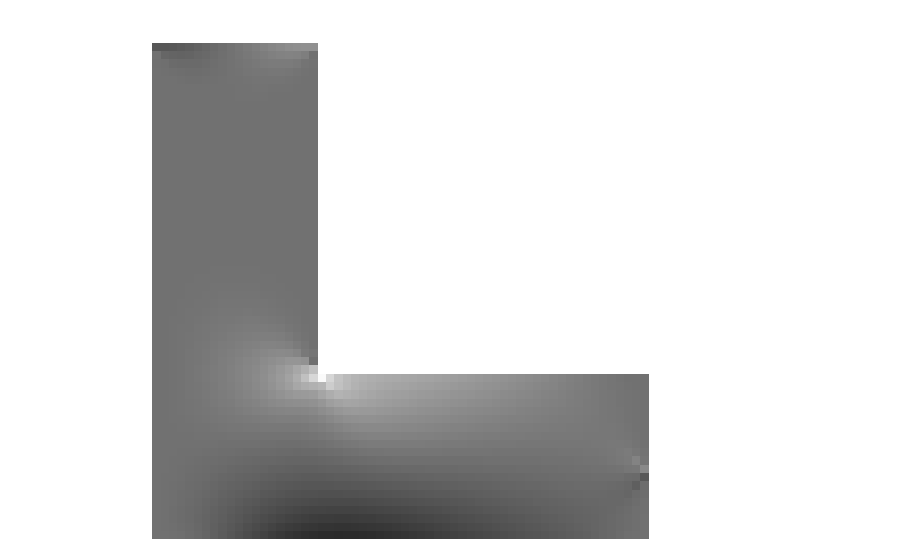} & \includegraphics[width=0.5\linewidth]{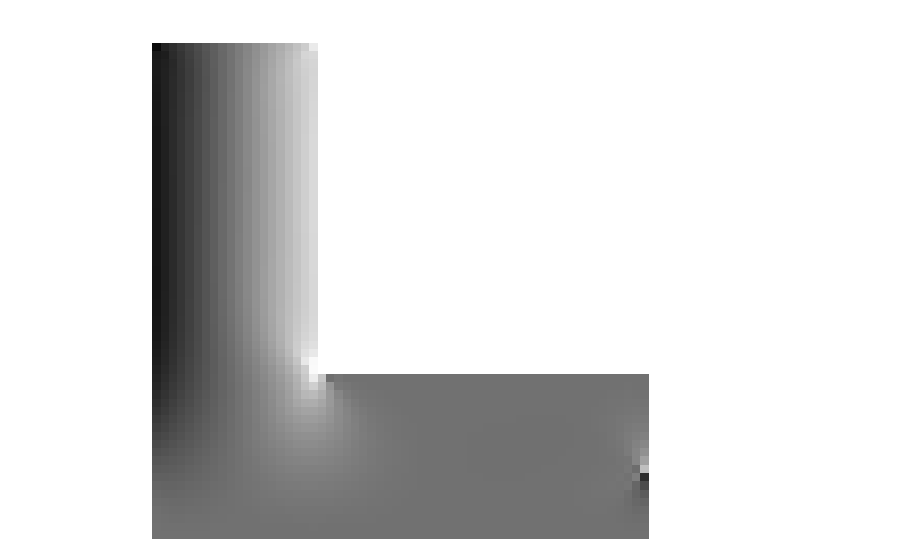} & 
			\includegraphics[width=0.5\linewidth]{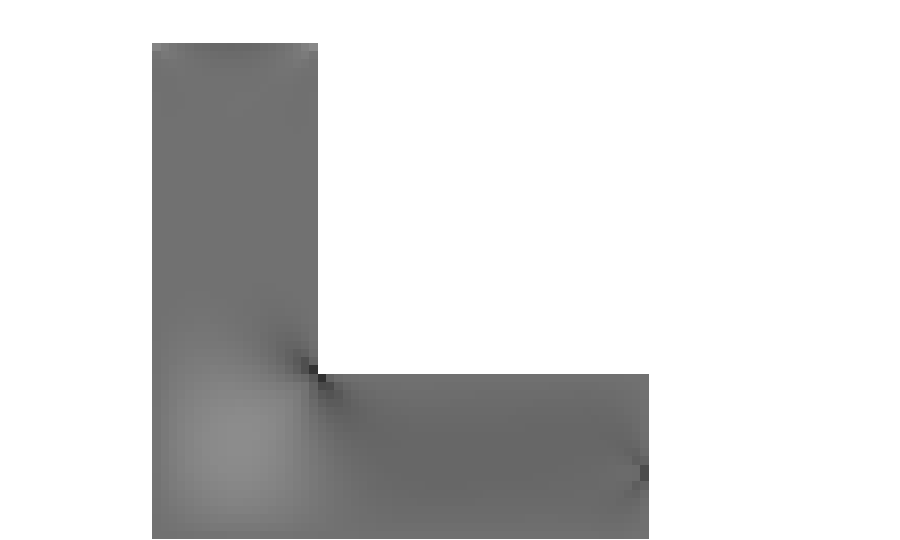}  \nextRow
			\hline
		\end{tabular}
	\end{center}
	\caption{L-shape example}
	\label{fig:L_ex}
\end{figure}

The six models proposed in section \ref{model} are trained using the same training dataset. 
The results of cantilever and L-shape examples are shown in \reffig{fig:bar_ex} and \reffig{fig:L_ex}, respectively. 
The super-resolved images of ESRGAN and PI-ESRGAN in \reffig{fig:bar_ex} and \reffig{fig:L_ex}, respectively, show that both models failed to super-resolve images.  

The loss values of U-Net-based models are listed in \reftab{tab:result}, where in the U-Net and UNet++ models, physical loss is not included; however, it is calculated to compare the results. 
U-Net exhibits the smallest MSE loss among all models, whereas PI-UNet exhibits the smallest physical loss. 
The PI-UNet reduced the physical loss at the cost of MSE loss, which led to a lower total loss. 
The loss values of UNet++ and PI-UNet++ were worse than those of UNet and PI-UNet, respectively. 
The network architecture of UNet++ is complicated compared with UNet. Hence, the number of network parameters to be trained was larger in UNet++ than in UNet. This difference causes a slight difference in the loss values. 

\begin{figure}[h]
	\begin{center}
		\begin{minipage}[h]{0.5\textwidth}
			\begin{center}
				\includegraphics[width=\linewidth]{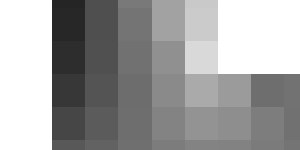}
				\par {(a) Coarse.}
			\end{center}
		\end{minipage}%
		\begin{minipage}[h]{0.5\textwidth}
			\begin{center}
				\includegraphics[width=\linewidth]{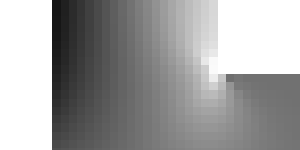}
				\par  {(b) Fine.}
			\end{center}
		\end{minipage}%
		\par
		\begin{minipage}[h]{0.5\textwidth}
			\begin{center}
				\includegraphics[width=\linewidth]{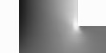}
				\par	{(c) U-Net.}
			\end{center}
		\end{minipage}%
		\begin{minipage}[h]{0.5\textwidth}
			\begin{center}
				\includegraphics[width=\linewidth]{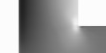}
				\par	{(d) PI-UNet.}
			\end{center}
		\end{minipage}%
		\caption{Close up view of coarse, fine, and super-resolved images.}
		\label{fig:closeup}
	\end{center}
\end{figure}

The close-up  view of a PI-UNet result is shown in \reffig{fig:closeup}. 
The mesh shape can be seen in the coarse and fine mesh, but the image becomes smooth in the UNet and PI-UNet results.  

The difference between ESRGAN-based  and UNet-based models arises from the difference in loss functions; the UNet-based models contain MSE loss, whereas the ESRGAN-based models do not. 
The ESRGAN-based models consider perceptual loss, and the generator is trained in an adversarial manner  with the discriminator. 
Hence, the discriminator has to be well trained to obtain a good generator. 
The generator does not know the ground truth; hence it is a difficult task for the generator to learn. 
For example, the super-resolved images of $\sigma_x$ in ESRGAN and $\sigma_y$ in PI-ESRGAN in \reffig{fig:bar_ex} are completely different from ground truth images. 
The generators output such wrong images because they do not know the ground truth. 
This learning instability is often observed in the GAN literature. 
In contrast, the loss functions of UNet-based models contain MSE loss between output and ground truth. Hence, the models can output at least similar data to the ground truth. The super-resolved images in \reffig{fig:bar_ex} and \reffig{fig:L_ex} qualitatively look the same as ground truth images. 
Consequently, its loss function and simple architecture are the reasons for the superiority of U-Net and PI-UNet. 

\begin{table}[htpb]
	\begin{center}
		\caption{Loss values of UNet-based models}
		\label{tab:result}       
		{\footnotesize 
			\begin{tabular}{l|rrr|rrr}
				\hline
				& \multicolumn{3}{c|}{Training loss} & \multicolumn{3}{c}{Validation loss} \\
				& Total loss & MSE loss & physical loss & Total loss & MSE loss & physical loss \\
				\hline
				UNet & 14.58$\times 10^{-4}$ & \textbf{1.18$\times 10^{-4}$} & 13.40$\times 10^{-4}$ & 19.83$\times 10^{-4}$ & \textbf{6.53$\times 10^{-4}$} & 13.28$\times 10^{-4}$ \\
				PI-UNet & \textbf{8.63$\times 10^{-4}$} & 4.44$\times 10^{-4}$ & \textbf{4.19$\times 10^{-4}$} & \textbf{13.80$\times 10^{-4}$} & 9.58$\times 10^{-4}$ & \textbf{4.22$\times 10^{-4}$} \\
				UNet++ & 22.90$\times 10^{-4}$ & 10.45$\times 10^{-4}$ & 12.45$\times 10^{-4}$ & 21.79$\times 10^{-4}$ & 8.99$\times 10^{-4}$ & 12.80$\times 10^{-4}$ \\
				PI-UNet++ & 18.92$\times 10^{-4}$ & 14.55$\times 10^{-4}$ & 4.37$\times 10^{-4}$ & 19.57$\times 10^{-4}$ & 15.17$\times 10^{-4}$ & 4.40$\times 10^{-4}$ \\
				\hline
			\end{tabular}%
		}
	\end{center}%
\end{table}%

\subsection{Generalization ability to unseen tasks}
\begin{figure}[h]
	\begin{center}
		\footnotesize
		\begin{tabular}{|C{5cm}{}{m}|C{5cm}{}{t}|C{5cm}{}{t}|C{5cm}{}{t}|}
			\hline
			Model & $\sigma_x$ & $\sigma_y$ & $\tau_{xy}$ \nextRow
			\hline 
			Coarse & \includegraphics[width=0.5\linewidth]{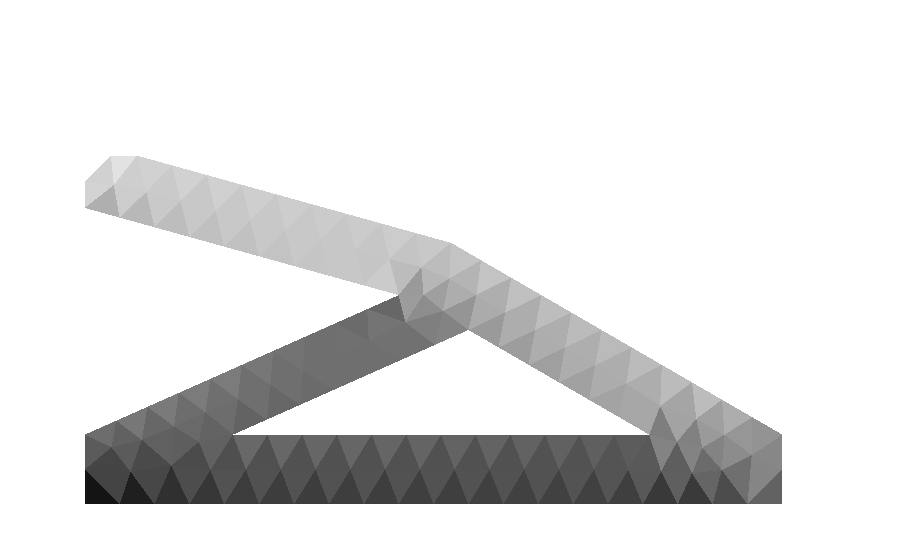} & \includegraphics[width=0.5\linewidth]{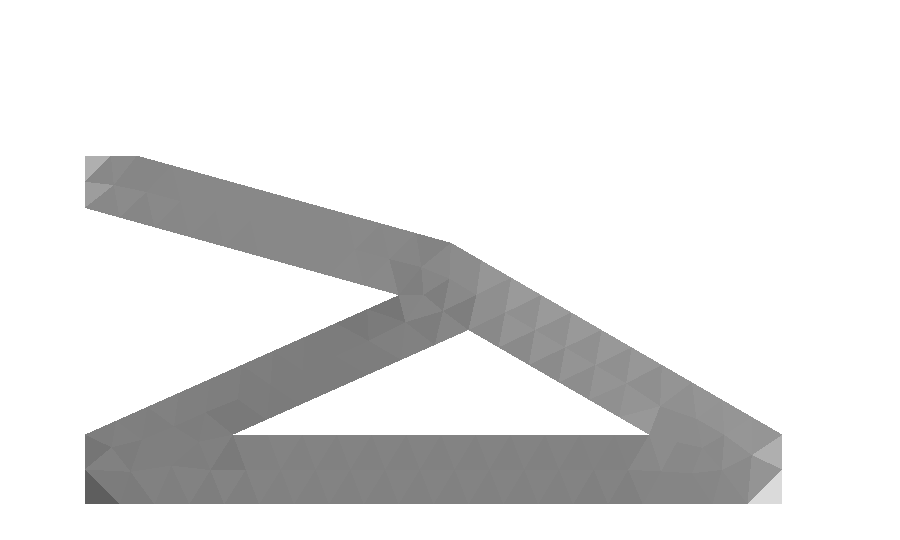} & 
			\includegraphics[width=0.5\linewidth]{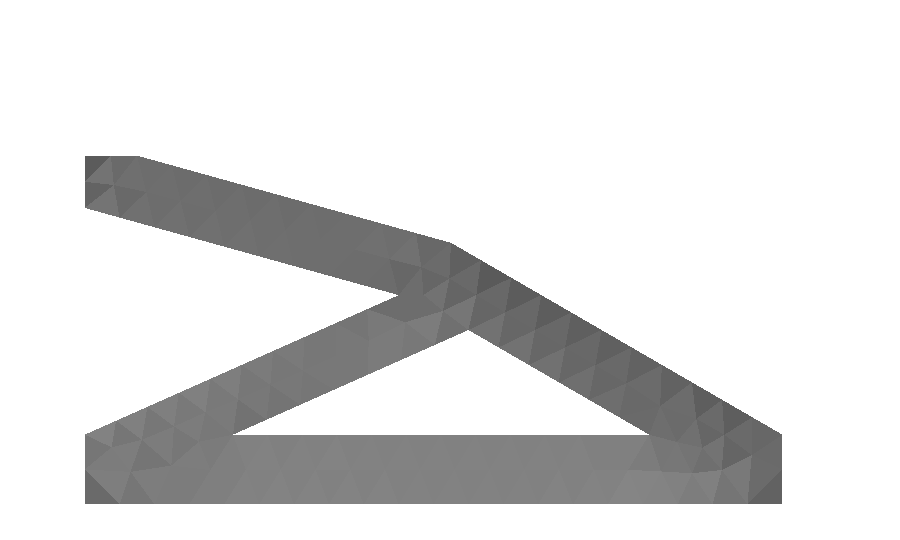}  \nextRow
			\hline 
			ESRGAN & \includegraphics[width=0.5\linewidth]{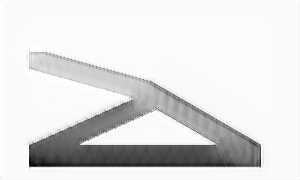} & \includegraphics[width=0.5\linewidth]{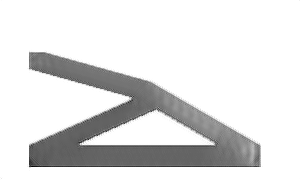} & \includegraphics[width=0.5\linewidth]{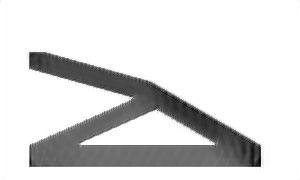}  \nextRow
			\hline
			PI-ESRGAN & \includegraphics[width=0.5\linewidth]{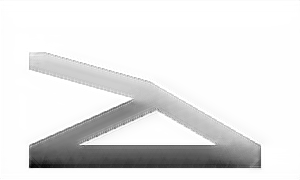} & \includegraphics[width=0.5\linewidth]{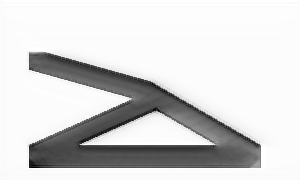} & \includegraphics[width=0.5\linewidth]{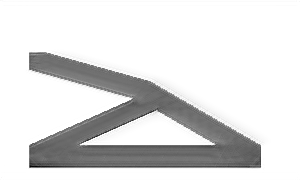}  \nextRow
			\hline
			U-Net & \includegraphics[width=0.5\linewidth]{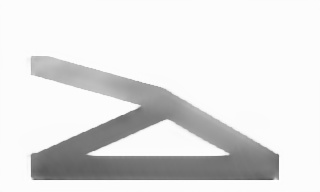} & \includegraphics[width=0.5\linewidth]{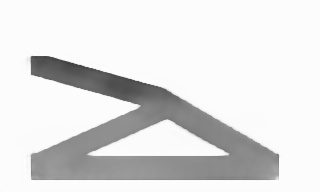} & \includegraphics[width=0.5\linewidth]{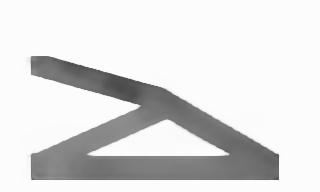}  \nextRow
			\hline
			PI-UNet & \includegraphics[width=0.5\linewidth]{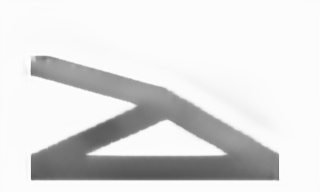} & \includegraphics[width=0.5\linewidth]{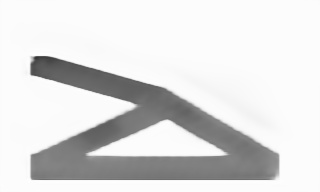} & \includegraphics[width=0.5\linewidth]{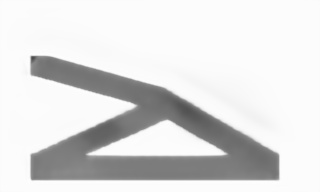}  \nextRow
			\hline
			U-Net++ & \includegraphics[width=0.5\linewidth]{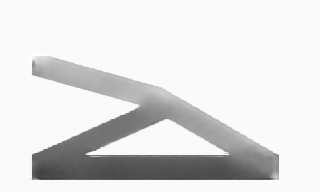} & \includegraphics[width=0.5\linewidth]{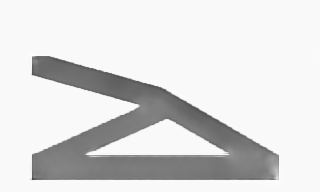} & \includegraphics[width=0.5\linewidth]{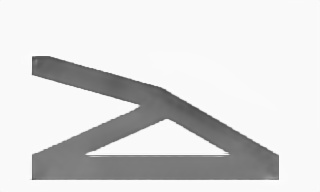}  \nextRow
			\hline
			PI-UNet++ & \includegraphics[width=0.5\linewidth]{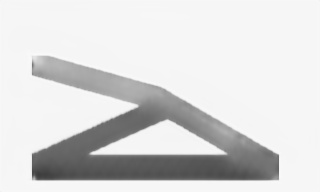} & \includegraphics[width=0.5\linewidth]{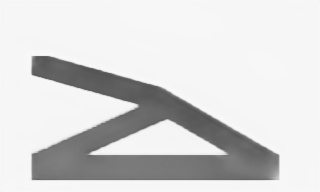} & \includegraphics[width=0.5\linewidth]{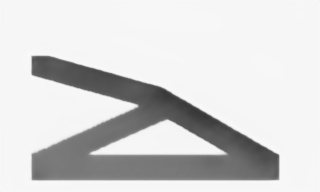}  \nextRow
			\hline
			Fine & \includegraphics[width=0.5\linewidth]{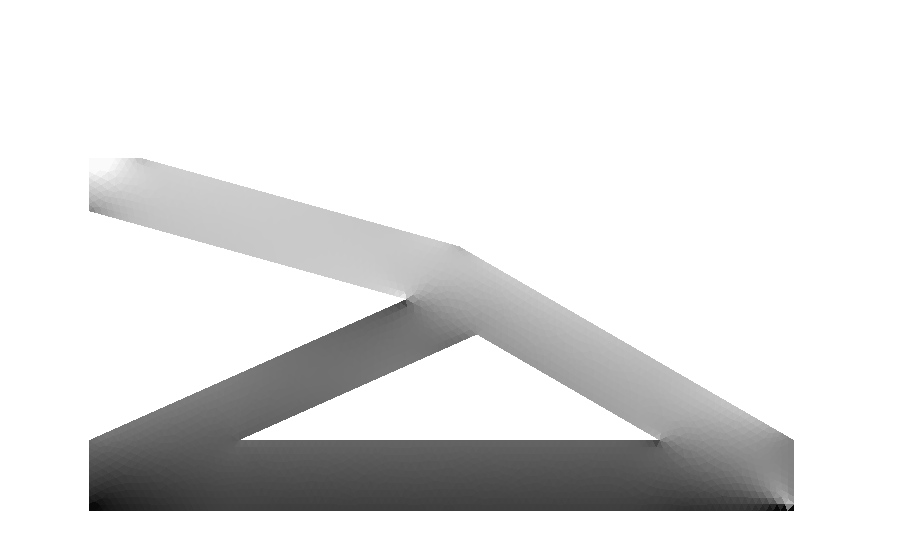} & \includegraphics[width=0.5\linewidth]{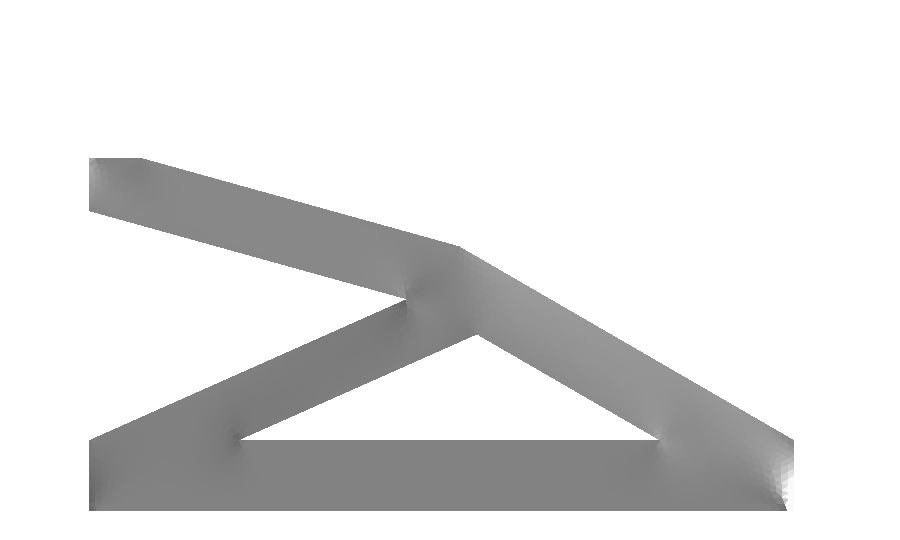} & 
			\includegraphics[width=0.5\linewidth]{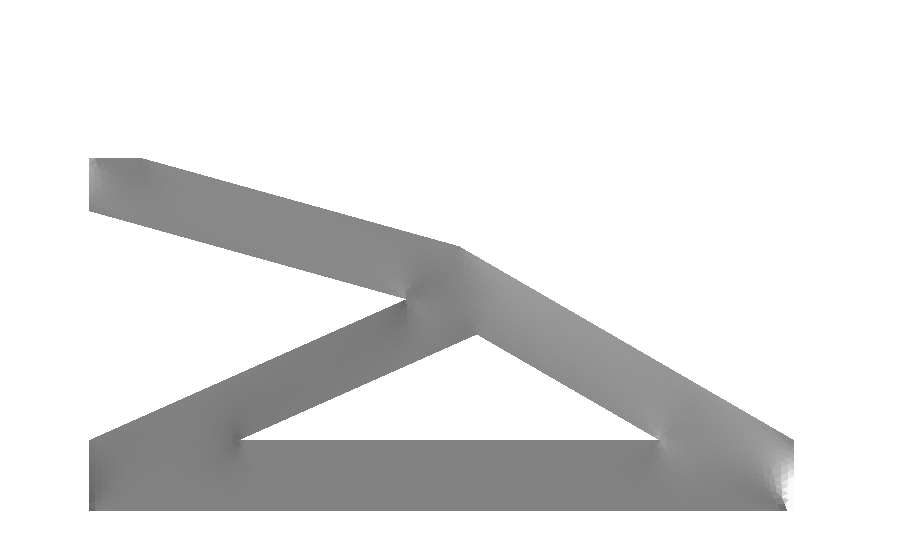}  \nextRow
			\hline
		\end{tabular}
	\end{center}
	\caption{Truss-like cantilever example}
	\label{fig:top}
\end{figure}

The trained model is validated using the truss-like cantilever. 
The losses are shown in \reftab{tab:unseen}, and the super-resolved images are shown in \reffig{fig:top}. All the super-resolved images reproduce the shape of the original truss-like cantilever, although the models are not trained with the data. The model is trained to draw contours only in the contour area of the input coarse mesh. 
However, ESRGAN-based models failed to predict accurately. The MSE  and physical losses of both models are greater than those of coarse mesh. 
Contrastingly , UNet-based models succeeded in super-resolving data.  In particular, the physical loss of PI-UNet is small. 
The PI-UNet can predict unfamiliar data from these experiments, although similar data is excluded from the training data.  
However, the prediction accuracy is worse than the experiments in section 4.1. It  is reasonable because, in section 4.1, the test data and training data are similar. 
Consequently, the model can be applied to unseen data; moreover, to improve accuracy, similar data should be added to the training dataset. 

\begin{table}[htpb]
	\begin{center}
		\caption{Loss values of an unseen task.}
		\label{tab:unseen}       
		{\footnotesize 
			\begin{tabular}{l|rrr}
				\hline
				& Total loss & MSE loss & physical loss \\
				\hline
				Coarse & 81.28$\times 10^{-4}$ & 76.74$\times 10^{-4}$ & 4.54$\times 10^{-4}$  \\
				ESRGAN & 133.09$\times 10^{-4}$ & 128.20$\times 10^{-4}$ & 4.89$\times 10^{-4}$  \\
				PI-ESRGAN & 117.30$\times 10^{-4}$ & 111.51$\times 10^{-4}$ & 5.79$\times 10^{-4}$  \\
				U-Net & 61.98$\times 10^{-4}$ & 57.75$\times 10^{-4}$ & 4.23$\times 10^{-4}$  \\
				PI-UNet & \textbf{60.14$\times 10^{-4}$} & 58.38$\times 10^{-4}$ & \textbf{1.76$\times 10^{-4}$}  \\
				UNet++ & 62.52$\times 10^{-4}$ & \textbf{57.72$\times 10^{-4}$} & 4.81$\times 10^{-4}$ \\
				PI-UNet++ & 83.50$\times 10^{-4}$ & 81.17$\times 10^{-4}$ & 2.33$\times 10^{-4}$  \\
				\hline
			\end{tabular}%
		}
	\end{center}%
\end{table}%

\section{Conclusion }
In this study, the UNet-based super-resolution model was proposed for predicting the stress tensor components from coarse mesh results.  
Because the equilibrium equation had to be satisfied in the FEM solution, physical loss was defined using divergence of the stress tensor.
The PI-UNet model indicated the best score while considering the MSE  and physics loss. 
Although ESRGAN is an outstanding model for super-resolution in image processing literature, the UNet-based models outperformed the ESRGAN-based models for the present task. 
The proposed model was validated using two types of data: test and validation data. The test data was similar to the training data. 
The model showed good agreement with the ground truth for the test data. Moreover, although the mesh shape was visible in ground truth data, it was not apparent in the super-resolved images, which is desirable from an application point of view. 
Contrastingly, the validation data is completely different unseen shapes. The model outputs a reasonable solution for such data. 
In particular, the physical loss is as small as that of the test data. 

The machine learning model should be trained with a sufficient amount  of data. However, in the FEM analysis application, it is not possible to collect all the possible data. This study shows that once trained, the proposed model is valid for data similar to the training data and outputs reasonable solutions for unseen tasks.

\section*{Acknowledgement}
This work was partially supported by JSPS KAKENHI Grant number 21K14064.

	\bibliographystyle{plain}
	\bibliography{bib-topologyopt.bib,bib-AM.bib,bib-DDD.bib}

\begin{thebibliography}{10}

\bibitem{Aage17}
Niels Aage, Erik Andreassen, Boyan~Stefanov Lazarov, and Ole Sigmund.
\newblock Giga-voxel computational morphogenesis for structural design.
\newblock {\em Nature}, 550:84--86, 2017.

\bibitem{Fleischmann19}
Christopher Fleischmann, Irina Leher, Reinhold Hartwich, Marc Hainke, and
  Stefan Sesselmann.
\newblock A new approach to quickly edit geometries and estimate stresses and
  displacements of implants in real-time.
\newblock {\em Current Directions in Biomedical Engineering}, 5:553--556, 2019.

\bibitem{Goodfellow14}
Ian~J. Goodfellow, Jean Pouget-Abadie, Mehdi Mirza, Bing Xu, David
  Warde-Farley, Sherjil Ozair, Aaron Courville, and Yoshua Bengio.
\newblock Generative adversarial nets.
\newblock In {\em Proceedings of the 27th International Conference on Neural
  Information Processing Systems - Volume 2}, NIPS'14, pages 2672--2680,
  Cambridge, MA, USA, 2014. MIT Press.

\bibitem{Jia21}
Xiaowei Jia, Jared Willard, Anuj Karpatne, Jordan~S. Read, Jacob~A. Zwart,
  Michael Steinbach, and Vipin Kumar.
\newblock Physics-guided machine learning for scientific discovery: An
  application in simulating lake temperature profiles.
\newblock {\em ACM/IMS Transactions on Data Science}, 2(3), may 2021.

\bibitem{Jiang21}
Haoliang Jiang, Zhenguo Nie, Roselyn Yeo, Amir~Barati Farimani, and
  Levent~Burak Kara.
\newblock Stress{GAN}: A generative deep learning model for {2D} stress
  distribution prediction.
\newblock {\em Journal of Applied Mechanics}, 88:1--11, 2021.

\bibitem{Alexia19}
Alexia Jolicoeur-Martineau.
\newblock The relativistic discriminator: a key element missing from standard
  {GAN}, 2019.

\bibitem{PINN}
George~Em Karniadakis, Ioannis~G. Kevrekidis, Lu~Lu, Paris Perdikaris, Sifan
  Wang, and Liu Yang.
\newblock Physics-informed machine learning.
\newblock {\em Nature Reviews Physics}, 3:422--440, 2021.

\bibitem{KUMAR17}
Mukesh Kumar, Trond Kvamsdal, and Kjetil~André Johannessen.
\newblock Superconvergent patch recovery and a posteriori error estimation
  technique in adaptive isogeometric analysis.
\newblock {\em Computer Methods in Applied Mechanics and Engineering},
  316:1086--1156, 2017.

\bibitem{Ledig17}
Christian Ledig, Lucas Theis, Ferenc Huszar, Jose Caballero, Andrew Cunningham,
  Alejandro Acosta, Andrew Aitken, Alykhan Tejani, Johannes Totz, Zehan Wang,
  and Wenzhe Shi.
\newblock Photo-realistic single image super-resolution using a generative
  adversarial network.
\newblock In {\em Proceedings of the IEEE Conference on Computer Vision and
  Pattern Recognition (CVPR)}, July 2017.

\bibitem{Lei19}
Hongshuai Lei, Chuanlei Li, Jinxin Meng, Hao Zhou, Yabo Liu, Xiaoyu Zhang,
  Panding Wang, and Daining Fang.
\newblock Evaluation of compressive properties of {SLM}-fabricated multi-layer
  lattice structures by experimental test and $\mu$-{CT}-based finite element
  analysis.
\newblock {\em Materials \& Design}, 169:107685, 2019.

\bibitem{Marinkovic19}
Dragan Marinkovic and Manfred Zehn.
\newblock Survey of finite element method-based real-time simulations.
\newblock {\em Applied sciences}, 9:2775, 2019.

\bibitem{Mukherjee21}
Sougata Mukherjee, Dongcheng Lu, Balaji Raghavan, Piotr Breitkopf, Subhrajit
  Dutta, Manyu Xiao, and Weihong Zhang.
\newblock Accelerating large-scale topology optimization: State-of-the-art and
  challenges.
\newblock {\em Archives of Computational Methods in Engineering},
  28:4549--4571, 2021.

\bibitem{Nakazawa19}
Shinji Nakazawa, Naoki Iwasaki, Ryuichi Matsuki, Gaku Hashimoto, Hiroshi Okuda,
  and Kazuya Goto.
\newblock High-resolution structural analysis of multilayer package substrate
  with open-source parallel {FE} software {F}ront{ISTR}.
\newblock In {\em 2019 IEEE CPMT Symposium Japan (ICSJ)}, pages 95--98, 2019.

\bibitem{UNet}
Olaf Ronneberger, Philipp Fischer, and Thomas Brox.
\newblock U-{N}et: Convolutional networks for biomedical image segmentation.
\newblock In Nassir Navab, Joachim Hornegger, William~M. Wells, and
  Alejandro~F. Frangi, editors, {\em Medical Image Computing and
  Computer-Assisted Intervention -- MICCAI 2015}, pages 234--241, Cham, 2015.
  Springer International Publishing.

\bibitem{VGG19}
Karen Simonyan and Andrew Zisserman.
\newblock Very deep convolutional networks for large-scale image recognition.
\newblock In Yoshua Bengio and Yann LeCun, editors, {\em 3rd International
  Conference on Learning Representations, {ICLR} 2015}, 2015.

\bibitem{Tan21}
Ren~Kai Tan, Chao Qian, Michael Wang, and Wenjing Ye.
\newblock An efficient data generation method for {ANN}-based surrogate models.
\newblock {\em Structural and Multidisciplinary Optimization}, 65:90, 2022.

\bibitem{Wang18}
Xintao Wang, Ke~Yu, Shixiang Wu, Jinjin Gu, Yihao Liu, Chao Dong, Yu~Qiao, and
  Chen Change~Loy.
\newblock {ESRGAN}: Enhanced super-resolution generative adversarial networks.
\newblock In {\em Proceedings of the European Conference on Computer Vision
  (ECCV) Workshops}, September 2018.

\bibitem{RankSRGAN}
Zhang Wenlong, Liu Yihao, Chao Dong, and Yu~Qiao.
\newblock Rank{SRGAN}: Generative adversarial networks with ranker for image
  super-resolution.
\newblock {\em IEEE Transactions on Pattern Analysis and Machine Intelligence},
  pages 1--1, 2021.

\bibitem{Willard20}
Jared Willard, Xiaowei Jia, Shaoming Xu, Michael Steinbach, and Vipin Kumar.
\newblock Integrating physics-based modeling with machine learning: A survey,
  2020.

\bibitem{Yadav14}
Praveen Yadav and Krishnan Suresh.
\newblock Large scale finite element analysis via assembly-free deflated
  conjugate gradient.
\newblock {\em Journal of Computing and Information Science in Engineering},
  14(4):041008, 2014.

\bibitem{Zhang20}
Ruiyang Zhang, Yang Liu, and Hao Sun.
\newblock Physics-guided convolutional neural network (phycnn) for data-driven
  seismic response modeling.
\newblock {\em Engineering Structures}, 215:110704, 2020.

\bibitem{UNet++}
Zongwei Zhou, Md~Mahfuzur~Rahman Siddiquee, Nima Tajbakhsh, and Jianming Liang.
\newblock {UNet++}: A nested {U-Net} architecture for medical image
  segmentation.
\newblock In {\em Deep Learning in Medical Image Analysis and Multimodal
  Learning for Clinical Decision Support, DLMIA 2018, ML-CDS 2018. Lecture
  Notes in Computer Science, Vol. 11045}, pages 3--11. Springer, Cham, 2018.

\bibitem{Zienkiewicz92a}
O.~C. Zienkiewicz and J.~Z. Zhu.
\newblock The superconvergent patch recovery and a posteriori error estimates.
  {P}art 1: The recovery technique.
\newblock {\em International Journal for Numerical Methods in Engineering},
  33(7):1331--1364, 1992.

\end{thebibliography}






\end{document}